\documentclass[letterpaper]{IEEEtran}
\usepackage{amsmath,amsfonts}
\usepackage{algorithmic}
\usepackage{array}
\usepackage[caption=false,font=normalsize,labelfont=sf,textfont=sf]{subfig}
\usepackage{textcomp}
\usepackage{stfloats}
\usepackage{url}
\usepackage{verbatim}
\usepackage{graphicx}
\usepackage{multirow}
\usepackage{xcolor}
\usepackage{arydshln}
\usepackage{pifont}
\newcommand{\cmark}{\textcolor{green!60!black}{\ding{51}}}
\newcommand{\xmark}{\textcolor{red!70!black}{\ding{55}}}
\usepackage{booktabs}
\usepackage{gensymb}
\usepackage{hyperref}
\usepackage{newtxtext,newtxmath} 

\def\BibTeX{{\rm B\kern-.05em{\sc i\kern-.025em b}\kern-.08em
    T\kern-.1667em\lower.7ex\hbox{E}\kern-.125emX}}
\usepackage{balance}

\begin{document}
\title{\emph{DetReIDX}: A Stress-Test Dataset for Real-World UAV-Based Person Recognition}

\author{Kailash A. Hambarde, Nzakiese Mbongo, Pavan Kumar MP, Satish Mekewad, Carolina Fernandes, Gökhan Silahtaroğlu, Alice Nithya, Pawan Wasnik, MD. Rashidunnabi, Pranita Samale, Hugo Proença~\IEEEmembership{Senior Member,~IEEE}%
\thanks{Manuscript received February XX, 2025; revised XX XX, 2025. This work was supported.}%
\thanks{Kailash A. Hambarde, Nzakiese Mbongo, Carolina Fernandes, MD. Rashidunnabi, Pranita Samale, and Hugo Proença are with the Instituto de Telecomunicações and the University of Beira Interior, Covilhã, Portugal (corresponding author e-mail: \href{mailto:kailas.srt@gmail.com}{kailas.srt@gmail.com}).}%
\thanks{Pavan Kumar MP is with J.N.N. College of Engineering, Shivamogga, Karnataka, India.}%
\thanks{Satish Mekewad and Pawan Wasnik are with the School of Computational Sciences, SRTM University, Nanded, India.}%
\thanks{Gökhan Silahtaroğlu is with Istanbul Medipol University, Istanbul, Turkey.}%
\thanks{Alice Nithya is with SRM Institute of Science and Technology, Kattankulathur, India.}%
}

\markboth{Submitted to IEEE Transactions on Biometrics, Behavior, and Identity Science}%
{Costa \MakeLowercase{\textit{et al.}}: \emph{DetReIDX}: A Stress-Test Dataset for Real-World UAV-Based Person Recognition}
\IEEEpubid{0000--0000/00\$00.00~\copyright~2021 IEEE}

\maketitle

\begin{abstract}
   Person reidentification (ReID) technology has been considered to perform relatively well under controlled, ground-level conditions, but it breaks down when deployed in challenging \emph{real-world} settings. Evidently, this is due to extreme data variability factors such as resolution, viewpoint changes, scale variations, occlusions, and appearance shifts from clothing or session drifts. Moreover, the publicly available data sets do not realistically incorporate such kinds and magnitudes of variability, which limits the progress of this technology.
  This paper introduces \emph{DetReIDX}, a large-scale aerial-ground person dataset, that was explicitly designed as a stress test to ReID under real-world conditions. \emph{DetReIDX} is a multi-session set that includes over 13 million bounding boxes from 509 identities, collected in seven university campuses from three continents, with drone altitudes between 5.8 and 120 meters. More important, as a key novelty, \emph{DetReIDX} subjects were recorded in (at least) two sessions on different days, with changes in clothing, daylight and location, making it suitable to actually evaluate \emph{long-term} person ReID. Plus, data were annotated from 16 soft biometric attributes and multitask labels for detection, tracking, ReID, and action recognition.
  In order to provide empirical evidence of \emph{DetReIDX} usefulness, we considered the specific tasks of human detection and ReID, where SOTA methods catastrophically degrade performance (up to 80\% in detection accuracy and over 70\% in Rank-1 ReID) when exposed to \emph{DetReIDX}'s conditions. The dataset, annotations, and official evaluation protocols are publicly available at \url{https://www.it.ubi.pt/DetReIDX/}.
\end{abstract}


\section{Introduction}
\IEEEPARstart{P}{erson} centric visual understanding including detection, identification, tracking, and re-identification (ReID) is foundational to a wide range of critical applications such as surveillance, public safety, autonomous UAV patrolling, and search-and-rescue operations \cite{chen2021disentangle}\cite{hu2025personvit}\cite{hambarde2024image}. However, the deployment of such systems in unconstrained aerial-ground environments remains extremely limited. The core bottleneck is not model capacity but rather the lack of datasets that reflect the true operational complexity of drone-based surveillance: low resolution, cross-viewpoint domain gaps, long-range degradation, and appearance shifts due to clothing or occlusion.
Despite impressive progress in ground-level person ReID using datasets like Market-1501~\cite{zheng2015scalable}, CUHK03~\cite{li2014deepreid}, MARS~\cite{zheng2016mars}, DukeMTMC-ReID~\cite{wu2018exploit}, and LTCC~\cite{qian2020long}, these benchmarks are largely constrained to fixed-camera, close-range, lateral-view scenarios. While they have catalyzed algorithmic advances, they fail to capture the severe viewpoint and scale variations encountered in aerial settings.

\begin{figure}[t]
  \centering
  \includegraphics[width=\linewidth]{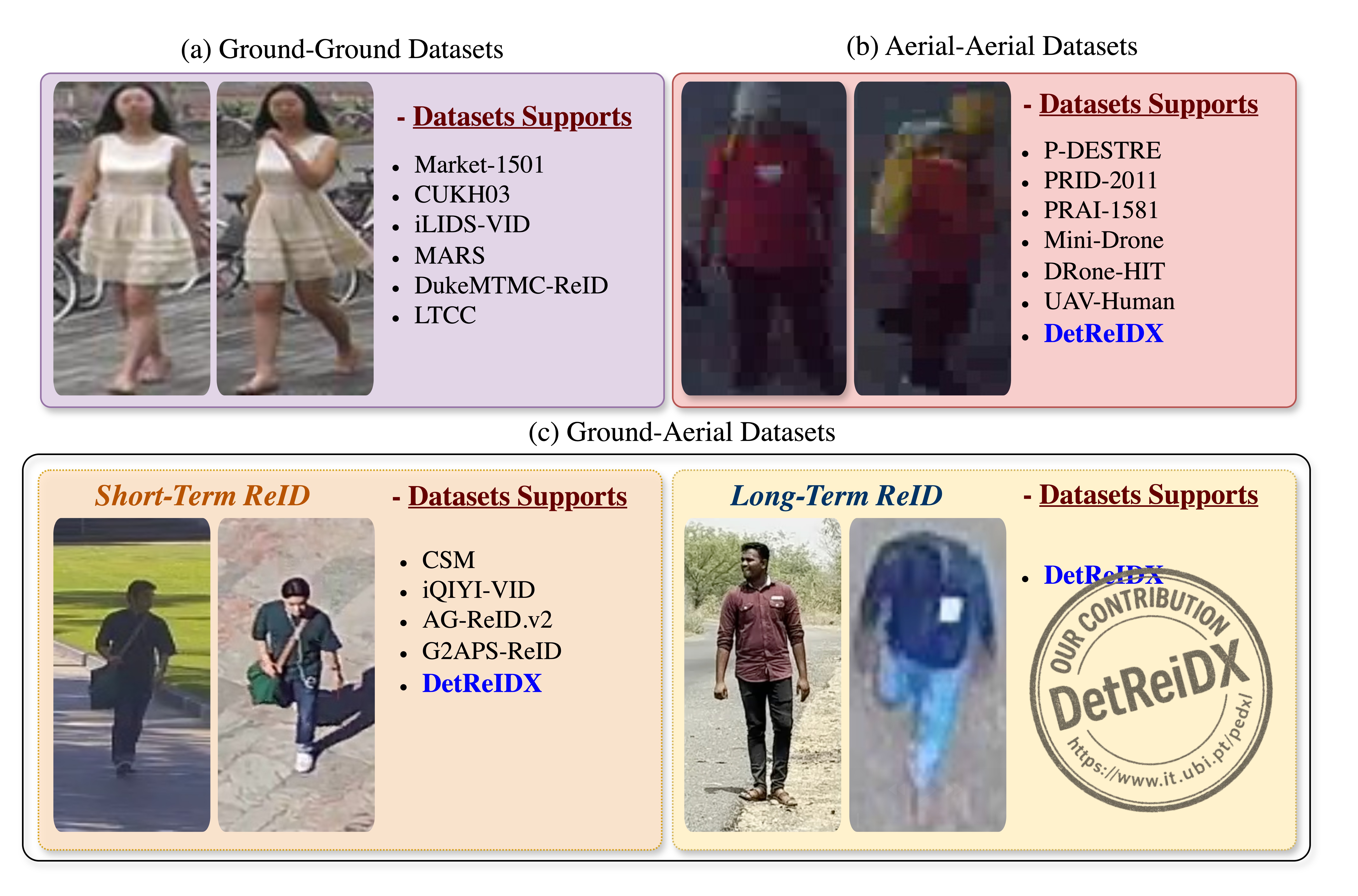}
  \caption{Comparison between the most important features of the publicly available datasets (ground-ground, aerial-aerial, and aerial-ground) and the \emph{DetReIDX} dataset. Unlike its counterparts, \emph{DetReIDX} includes clothing variations \emph{within subjects}, with detection and tracking annotations, action labels, at wide altitude ranges (5.8m–120m).}
  \label{fig:abstractfig}
\end{figure}

\begin{table*}[ht]
  \centering
  \caption{Comparison between \emph{DetReIDX} and the publicly available datasets for person detection, ReID, tracking, and action recognition. (\cmark: Available, \xmark: Not available, --: No information available.)}
  \label{tab:dataset_comparison}
  \renewcommand{\arraystretch}{1.2}
  \resizebox{\textwidth}{!}{
  \begin{tabular}{ccccccccccccc}
    \hline
    \multirow{2}{*}{\textbf{Category}} &
    \multirow{2}{*}{\textbf{Dataset}} &
    \multirow{2}{*}{\textbf{Camera}} &
    \multirow{2}{*}{\textbf{Format}} &
    \multicolumn{5}{c}{\textbf{Task}} &
    \multirow{2}{*}{\textbf{\#Identities}} &
    \multirow{2}{*}{\textbf{\#BBox}} &
    \multirow{2}{*}{\textbf{Height (m)}} &
    \multirow{2}{*}{\textbf{Distance (m)}} \\
    
    \cline{5-9}
    & & & & \textbf{Detection} & \textbf{Tracking} & \textbf{ReID} & \textbf{Search} & \textbf{Action Rec.} & & & & \\
    \hline

    \multirow{6}{*}{\rotatebox[origin=c]{90}{\textbf{Ground-Ground}}}
    & CUHK03 \cite{li2014deepreid} & CCTV & Still & \xmark & \xmark & \cmark & \xmark & \xmark & 1467 & 13K & -- & -- \\
    & iLIDS-VID \cite{wang2014person} & CCTV & Video & \xmark & \xmark & \cmark & \xmark & \xmark & 300 & 42K & -- & -- \\
    & Market-1501 \cite{zheng2015scalable} & CCTV & Still & \cmark & \cmark & \cmark & \xmark & \xmark & 1501 & 32.6K & $<$10 & -- \\
    & MARS \cite{zheng2016mars} & CCTV & Video & \cmark & \cmark & \cmark & \xmark & \xmark & 1261 & 20K & -- & -- \\
    & DukeMTMC-ReID \cite{wu2018exploit} & CCTV & Video & \cmark & \cmark & \cmark & \xmark & \xmark & 1812 & 815K & -- & -- \\
    & LTCC \cite{qian2020long} & CCTV & Still & \cmark & \xmark & \cmark & \cmark & \xmark & 152 & 17K & -- & -- \\
    \hdashline

    \multirow{8}{*}{\rotatebox[origin=c]{90}{\textbf{Aerial-Aerial}}}
    & PRID-2011 \cite{hirzer2011person} & UAV & Still & \xmark & \xmark & \cmark & \xmark & \xmark & 1581 & 40K & 20--60 & -- \\
    & MRP \cite{layne2015investigating} & UAV & Video & \cmark & \cmark & \cmark & \xmark & \xmark & 28 & 4K & $<$10 & -- \\
    & PRAI-1581 \cite{zhang2020person} & UAV & Still & \xmark & \xmark & \cmark & \xmark & \xmark & 1581 & 39K & 20--60 & -- \\
    & Mini-drone \cite{bonetto2015privacy} & UAV & Video & \cmark & \cmark & \xmark & \xmark & \cmark & -- & $>$27K & $<$10 & -- \\
    & AVI \cite{singh2018eye} & UAV & Still & \cmark & \cmark & \cmark & \cmark & \cmark & 5124 & 10K & 2--8 & -- \\
    & DRone-HIT \cite{grigorev2019deep} & UAV & Still & \cmark & \xmark & \cmark & \cmark & \xmark & 101 & 40K & -- & -- \\
    & P-DESTRE \cite{kumar2020p} & UAV & Video & \cmark & \cmark & \cmark & \cmark & \cmark & 269 & $>$14.8M & 5.8--6.7 & -- \\
    & UAV-Human \cite{li2021uav} & UAV & Still & \xmark & \xmark & \cmark & \xmark & \xmark & 1144 & 41K & 2--8 & -- \\
    \hdashline

    \multirow{5}{*}{\rotatebox[origin=c]{90}{\textbf{Aerial-Ground}}}
    & CSM \cite{ahmed2015using} & Various & Video & \xmark & \xmark & \cmark & \xmark & \xmark & 1218 & 11M & -- & -- \\
    & iQIYI-VID \cite{liu2018iqiyi} & Various & Video & \cmark & \cmark & \cmark & \cmark & \xmark & 5000 & 600K & -- & -- \\
    & AG-ReID.v2 \cite{nguyen2024ag} & UAV+CCTV & Still & \cmark & \cmark & \cmark & \cmark & \xmark & 1615 & 100.6K & 15--45 & -- \\
    & G2APS-ReID \cite{li2021uav} & UAV+CCTV & Still & \cmark & \cmark & \cmark & \cmark & \xmark & 2788 & 200.8K & 20--60 & -- \\
    & \emph{DetReIDX} (Ours) & DSLR+UAV & Video+Still & \cmark & \cmark & \cmark & \cmark & \cmark & 509 & 12.6M & 5--120 & 10--120 \\
    \hline
  \end{tabular}}
\end{table*}

On the other hand, aerial-only datasets such as P-DESTRE~\cite{kumar2020p}, UAV-Human~\cite{li2021uav}, PRID-2011~\cite{hirzer2011person}, MRP~\cite{layne2015investigating}, PRAI-1581~\cite{zhang2020person}, Mini-drone~\cite{bonetto2015privacy}, AVI~\cite{singh2018eye}, and DRone-HIT~\cite{grigorev2019deep} offer aerial captures but are  limited to relatively low altitudes ($<$10m), lack multi-session diversity, or exclude ground-view perspectives, thus limiting their value for cross-view understanding and realistic tracking tasks.
Bridging the aerial-ground domain remains vastly underexplored. Notable attempts include AG-ReID.v2~\cite{nguyen2024ag}, G2APS~\cite{wang2025secap}, CSM~\cite{ahmed2015using}, and iQIYI-VID~\cite{liu2018iqiyi}, which introduce hybrid viewpoints. Yet, these datasets suffer from narrow altitude ranges (typically $<$45m), limited clothing variation, and lack fine-grained annotations necessary for robust multi-task learning.

\noindent\textbf{The gap:} Existing datasets either (i) operate in narrow altitude domains, (ii) fail to support cross-view matching, (iii) lack annotation density and appearance variation to evaluate long-term recognition, or (iv) omit long-term identity retention under clothing changes across sessions. Most benchmarks assume fixed attire and short-term reappearance, which breaks down in real-world scenarios where individuals are observed days apart in different clothing. This makes current benchmarks fundamentally unsuitable for training or stress-testing models intended for UAV-based deployments.

To address this, we propose \emph{DetReIDX}, a large-scale, aerial-ground person dataset specifically designed to evaluate model robustness under real-world constraints. \emph{DetReIDX} includes:
\begin{itemize}
    \item 13M+ bounding boxes from 509 subjects, recorded in 7 universities of 3 different continents (Portugal, Turkey, India and Angola).
    \item Data spanning 5.8m to 120m altitude and 10m to 120m distance, across 18 unique UAV viewpoints.
    \item Aerial, and ground views captured in two distinct sessions, to support clothing variation and temporal drift.
    \item Manual annotations of 16 soft biometric attributes~\cite{kumar2020p} (e.g., age, gender, height, hair style, upper/lower clothing, accessories).
    \item Multi-task labels for detection, ReID, action recognition, tracking, and cross-domain matching.
\end{itemize}

\vspace{1mm}
\noindent\textbf{Why \emph{DetReIDX} matters:} Figure~\ref{fig:abstractfig} and Table~\ref{tab:dataset_comparison} show that \emph{DetReIDX} dramatically exceeds previous datasets in altitude range, viewpoint coverage, identity diversity and annotation richness. In our experiments, SOTA detection models such as YOLOv8~\cite{roboflow2023yolov8}, DDOD~\cite{chen2021disentangle}, and Grid-RCNN~\cite{lu2019grid} degrade by up to 80\% when transferred to long-range (D3) scenes. Similarly, leading ReID methods including PersonViT~\cite{hu2025personvit}, SeCap~\cite{wang2025secap}, and CLIP-ReID~\cite{li2023clip} collapse when subject to aerial-ground viewpoint shifts and appearance changes. 

Crucially, \emph{DetReIDX} is the first to explicitly incorporate long-term identity variation via clothing changes across sessions, revealing how heavily current ReID models rely on superficial appearance cues rather than learning semantically grounded or structural identity features. This makes \emph{DetReIDX} not only harder, but closer to operational reality and indispensable for progress.

\vspace{1mm}
\noindent\textbf{Contributions:}
\begin{itemize}
    \item We announce and describe the \emph{DetReIDX} set, the most comprehensive person-centric dataset designed for UAV-ground multi-task benchmarking under real-world conditions.
    \item We provide empirical evidence about SOTA models failure to generalize under realistic and very challenging \emph{real-wordl} settings.
    \item We provide a rigorous set of benchmarks for detection and ReID tasks, highlighting the current imitations and pointing to new research directions for robust cross-view ReID.
\end{itemize}

The remainder of this paper is organized as follows: Section~\ref{sec:related} gives an overview of the related sets and the limitations of the existing benchmarks. Section~\ref{sec:dataset} details the data collection and annotation procedures. Section~\ref{sec:experiments} presents task-specific experiments and results. Finally, Section~\ref{sec:conclusion} concludes the paper.

\begin{table*}[ht]
  \centering
  \footnotesize
  \caption{Comparison between the available person annotations in the existing datasets. (\cmark\ stand for attribute available and \xmark\ indicate unavailability).}
  \label{tab:attribute_comparison}
  \resizebox{\textwidth}{!}{%
  \begin{tabular}{l cccc ccccccc cccccc}
    \toprule
    & \multicolumn{4}{c}{\textbf{Ground-Ground}} & \multicolumn{7}{c}{\textbf{Aerial-Aerial}} & \multicolumn{4}{c}{\textbf{Aerial-Ground}} \\
    \cmidrule(lr){2-5} \cmidrule(lr){6-12} \cmidrule(lr){13-17}
    \textbf{Attribute} & \textbf{Market-1501} & \textbf{DukeMTMC} & \textbf{CUHK03} & \textbf{iLIDS-VID} & 
    \textbf{P-DESTRE} & \textbf{UAV-Human} & \textbf{PRID-2011} & \textbf{MRP} & \textbf{PRAI-1581} & 
    \textbf{Mini-Drone} & \textbf{AVI} & \textbf{AG-ReID.v1} & \textbf{AG-ReID.v2} & \textbf{G2APS} & \textbf{iQIYI-VID} & \textbf{\emph{DetReIDX} (Our)}\\
    \midrule
    Gender             & \cmark & \cmark & \xmark & \xmark & \cmark & \cmark & \xmark & \xmark & \xmark & \xmark & \xmark & \cmark & \cmark & \xmark & \xmark & \cmark \\
    Age                & \cmark & \xmark & \xmark & \xmark & \cmark & \xmark & \xmark & \xmark & \xmark & \xmark & \xmark & \xmark & \cmark & \xmark & \xmark & \cmark \\
    Height             & \xmark & \xmark & \xmark & \xmark & \cmark & \xmark & \xmark & \xmark & \xmark & \xmark & \xmark & \xmark & \cmark & \xmark & \xmark & \cmark \\
    Body Volume        & \xmark & \xmark & \xmark & \xmark & \cmark & \xmark & \xmark & \xmark & \xmark & \xmark & \xmark & \xmark & \cmark & \xmark & \xmark & \cmark \\
    Ethnicity          & \xmark & \xmark & \xmark & \xmark & \cmark & \xmark & \xmark & \xmark & \xmark & \xmark & \xmark & \xmark & \cmark & \xmark & \xmark & \cmark \\
    Hair Color         & \xmark & \xmark & \xmark & \xmark & \cmark & \xmark & \xmark & \xmark & \xmark & \xmark & \xmark & \xmark & \cmark & \xmark & \xmark & \cmark \\
    Hairstyle          & \cmark & \xmark & \xmark & \xmark & \cmark & \xmark & \xmark & \xmark & \xmark & \xmark & \xmark & \xmark & \cmark & \xmark & \xmark & \cmark \\
    Beard              & \xmark & \xmark & \xmark & \xmark & \cmark & \xmark & \xmark & \xmark & \xmark & \xmark & \xmark & \xmark & \cmark & \xmark & \xmark & \cmark \\
    Moustache          & \xmark & \xmark & \xmark & \xmark & \cmark & \xmark & \xmark & \xmark & \xmark & \xmark & \xmark & \xmark & \cmark & \xmark & \xmark & \cmark \\
    Glasses            & \xmark & \xmark & \xmark & \xmark & \cmark & \xmark & \xmark & \xmark & \xmark & \xmark & \xmark & \cmark & \cmark & \xmark & \xmark & \cmark \\
    Head Accessories   & \cmark & \cmark & \xmark & \xmark & \cmark & \cmark & \xmark & \xmark & \xmark & \xmark & \xmark & \cmark & \cmark & \xmark & \xmark & \cmark \\
    Upper Body Clothing & \cmark & \cmark & \xmark & \xmark & \cmark & \cmark & \xmark & \xmark & \xmark & \xmark & \xmark & \cmark & \cmark & \xmark & \xmark & \cmark \\
    Lower Body Clothing & \cmark & \cmark & \xmark & \xmark & \cmark & \cmark & \xmark & \xmark & \xmark & \xmark & \xmark & \cmark & \cmark & \xmark & \xmark & \cmark \\
    Feet               & \xmark & \cmark & \xmark & \xmark & \cmark & \xmark & \xmark & \xmark & \xmark & \xmark & \xmark & \xmark & \cmark & \xmark & \xmark & \cmark \\
    Accessories        & \cmark & \cmark & \xmark & \xmark & \cmark & \cmark & \xmark & \xmark & \xmark & \xmark & \xmark & \cmark & \cmark & \xmark & \xmark & \cmark \\
    Action             & \xmark & \xmark & \xmark & \xmark & \cmark & \xmark & \xmark & \xmark & \xmark & \xmark & \cmark & \xmark & \xmark & \xmark & \xmark & \cmark \\
    \bottomrule
  \end{tabular}%
  }
\end{table*}

\begin{figure}[!t]
  \centering
  \includegraphics[width=0.95\columnwidth]{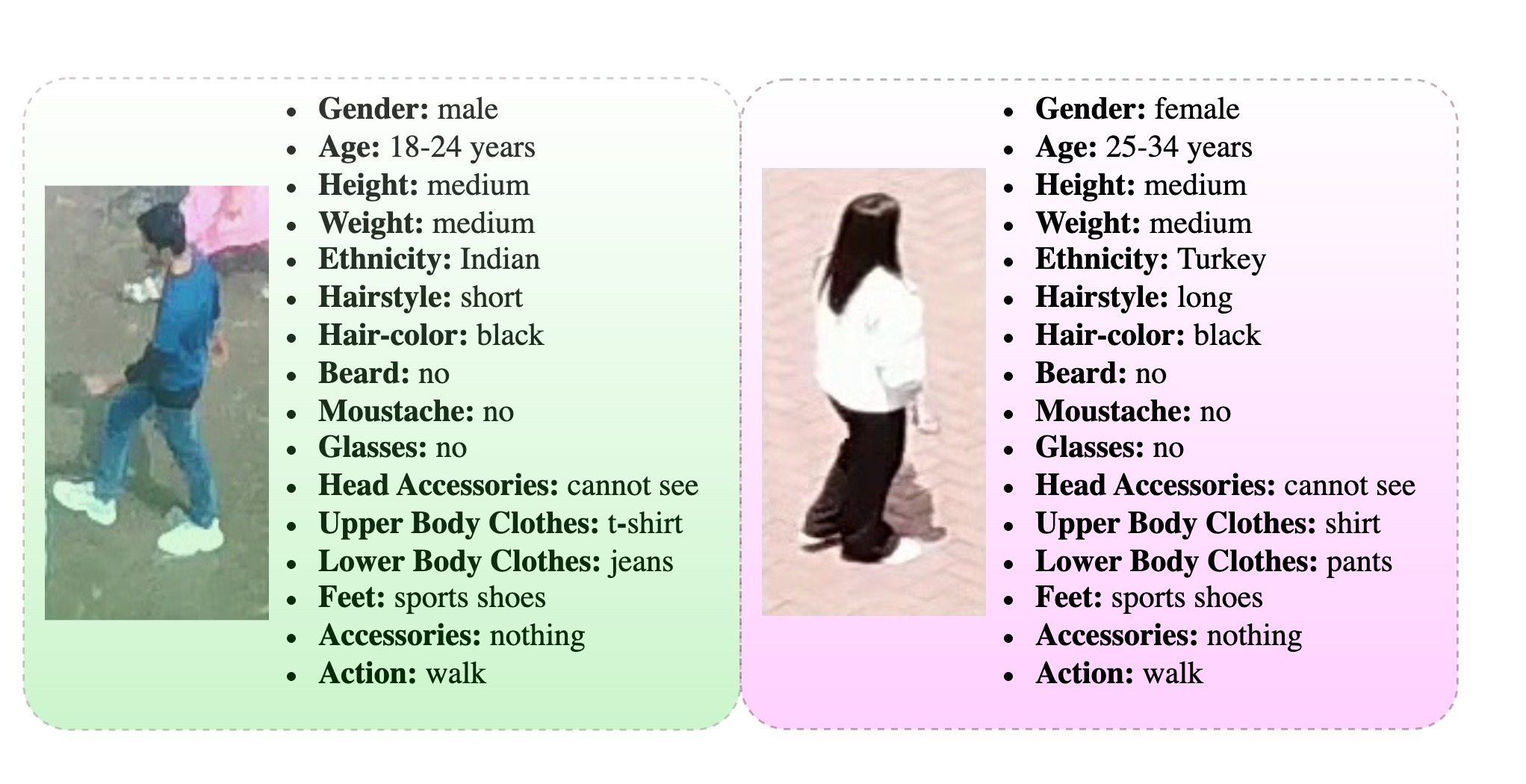}
  \caption{Examples of soft biometric annotations for two individuals in the \emph{DetReIDX} dataset. Each subject is labeled with 16 visual and demographic attributes, facilitating fine-grained person analysis across multiple scenes.}
  \label{fig:DetReIDX_soft_biometrics}
\end{figure}

\section{Related Work}
\label{sec:related}

Person recognition from visual data has been receiving growing attention by the reserch community. However, most of the existing datasets and benchmarks fall into three isolated silos \textit{ground-ground}, \textit{aerial-aerial}, or \textit{aerial-ground} each with critical limitations when viewed through the lens of UAV-based long-range surveillance.

\subsection{Ground-Ground Datasets}
Ground-level ReID datasets such as Market-1501~\cite{zheng2015scalable}, CUHK03~\cite{li2014deepreid}, MARS~\cite{zheng2016mars}, DukeMTMC-ReID~\cite{wu2018exploit}, and LTCC~\cite{qian2020long} have become standard testbeds for model development. These datasets enable benchmarking across appearance changes, occlusion, and temporal variations. However, all are collected from static ground cameras with minimal viewpoint variation and no aerial data. Crucially, subjects are captured at close range with full-body visibility conditions that are fundamentally different from long-range aerial footage. As a result, models trained on these datasets fail to generalize to UAV deployment scenarios.

\subsection{Aerial-Aerial Datasets}
Datasets like PRID-2011~\cite{hirzer2011person}, PRAI-1581~\cite{zhang2020person}, MRP~\cite{layne2015investigating}, Mini-drone~\cite{bonetto2015privacy}, and P-DESTRE~\cite{kumar2020p} shift focus to aerial-only captures. While they introduce novel challenges such as low resolution and top-down views, they suffer from two key limitations: 1) extremely low altitude ranges (typically under 10m), which do not reflect true UAV flight conditions; and 2) the absence of any ground perspective, making them unsuitable for cross-view ReID or domain-bridging tasks. Even advanced datasets like UAV-Human~\cite{li2021uav} and AVI~\cite{singh2018eye} lack consistent identity tracking across multiple angles and distances.

\subsection{Aerial-Ground Datasets}
A handful of datasets attempt to bridge the domain gap between UAV and CCTV cameras most notably AG-ReID.v2~\cite{nguyen2024ag}, G2APS~\cite{wang2025secap}, CSM~\cite{ahmed2015using}, and iQIYI-VID~\cite{liu2018iqiyi}. These efforts mark important progress but are fundamentally limited in scope: Their altitude range is narrow (typically 15--45\,m), excluding high-altitude drone perspectives. Clothing variation across sessions is minimal or absent, reducing the challenge of long-term ReID. Annotations are limited to ReID detection, tracking, action recognition, and soft biometrics are often missing. Cross-session and cross-location diversity is limited, reducing real-world generalization.

\subsection{Where \emph{DetReIDX} Fits}
Unlike all prior datasets, \emph{DetReIDX} is designed to address the realities of long-range, cross-domain person understanding:

\begin{itemize}
    \item \textbf{Altitude and Distance Diversity:} Captures span from 5.8\,m to 120\,m in altitude, and 10\,m to 120\,m in lateral range far beyond any existing benchmark.

    \item \textbf{Aerial-Ground Pairing:} Each subject is recorded in controlled indoor conditions (ground views) and from 18 aerial viewpoints, enabling rich cross-domain matching.

    \item \textbf{Session-Wise Clothing Variation:} Subjects are recorded across multiple days with different outfits. This explicitly simulates \textit{long-term ReID}, where appearance changes due to clothing occlude texture- and color-based identity cues. Unlike AG-ReID and G2APS, \emph{DetReIDX} exposes how fragile modern ReID systems are when color, clothing, or silhouette cannot be relied on.

    \item \textbf{Comprehensive Multi-Task Annotation:} In addition to ReID labels, \emph{DetReIDX} provides bounding boxes, tracking IDs, action labels, and 16 soft biometric attributes supporting detection, identification, and fine-grained analysis under extreme scale and occlusion conditions.
\end{itemize}

\noindent \textbf{Key distinction:} Where prior datasets isolate either viewpoint, task, or domain, \emph{DetReIDX} unifies them. It offers a systematic breakdown of how model performance degrades under scale shift, viewpoint change, occlusion, and appearance drift setting a new benchmark for \textit{aerial-to-ground person understanding under real-world constraints}.

\section{The \emph{DetReIDX} Dataset}
\label{sec:dataset}

\begin{figure}[t]
  \centering
  \includegraphics[width=\linewidth]{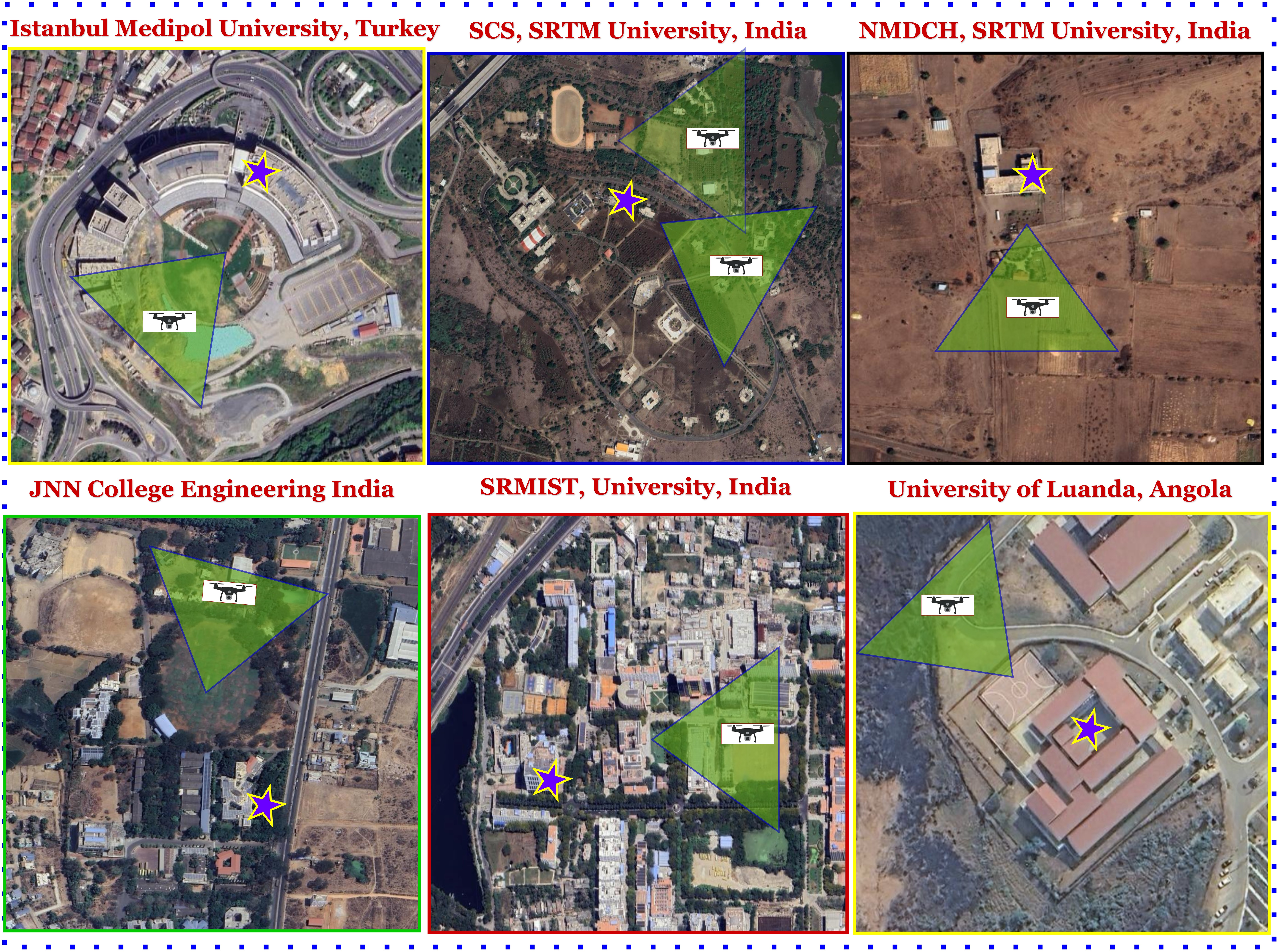}
  \caption{Satellite view of the data collection sites across the university campuses in Turkey, Angola, and India. The star markers indicate indoor dataset collection, and the green cones represent drone flight zones.}
  \label{fig:DetReIDX_maps}
\end{figure}

\emph{DetReIDX} is a comprehensive dataset for long-range, cross-view person understanding. It enables detection, tracking, identification, ReID, and soft-biometric prediction across aerial and ground views. \emph{DetReIDX} is built from the ground up to reflect real-world constraints faced by UAV surveillance: multi-view occlusion, top-down distortion, extreme resolution loss, appearance shifts, and domain gaps between aerial and ground captures.

The dataset includes over 13 million bounding boxes from 509 identities, with consistent ID annotation across two capture sessions and three continents. All participants are annotated with 16 soft biometric attributes and captured using a structured, hierarchical drone protocol to support controlled evaluation under varied pitch, altitude, and distance.

\subsection{Collection Sites and Demographic Diversity}

\emph{DetReIDX} was collected in seven universities from India, Portugal, Turkey, and Angola, as shown in Figure~\ref{fig:DetReIDX_maps}. The selection of geographically and culturally distinct campuses ensures diversity in subject appearance, environment, clothing, and lighting—enabling broader generalization.

In total, the dataset includes 509 subjects, each with indoor and outdoor recordings. Participants span across a wide range of height, weight, ethnicity, and other appearance attributes (see Figure~\ref{fig:DetReIDX_soft_biometrics}).


\begin{figure}[t]
  \centering
  \includegraphics[width=\linewidth]{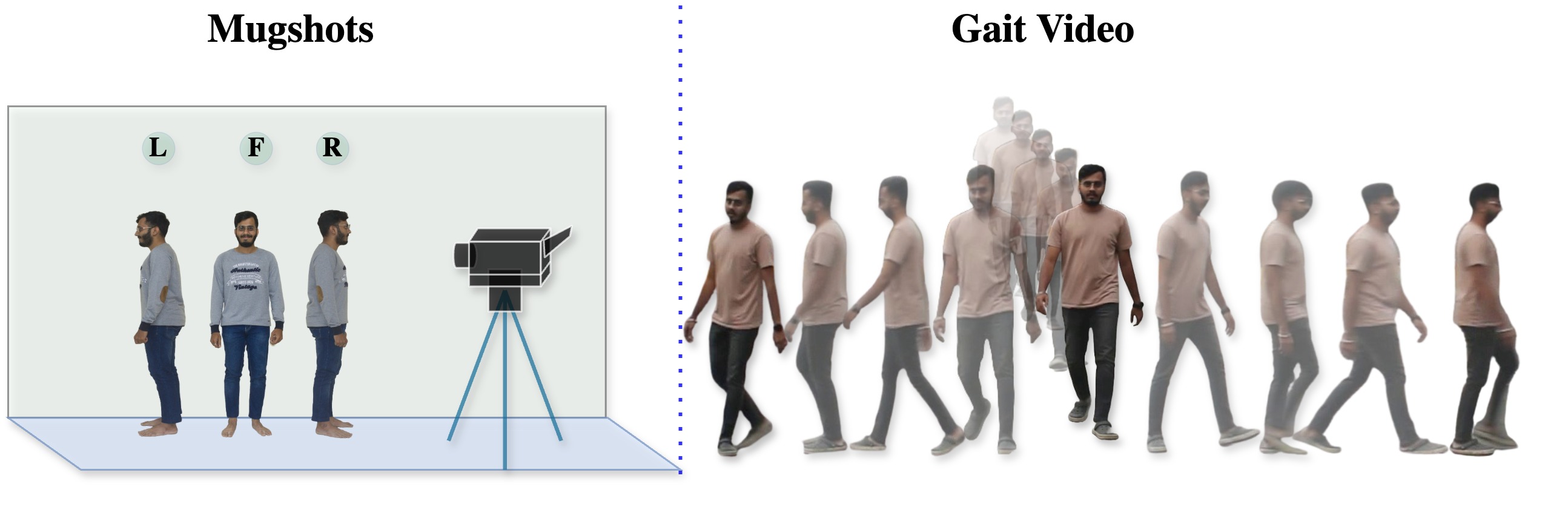}
  \caption{Overview of the indoor data collection setup: (left) mugshots taken from three angles (left, front, right); (right) gait video.}
  \label{fig:indoor_setup}
\end{figure}


\begin{figure}[ht]
  \centering
  \includegraphics[width=\linewidth]{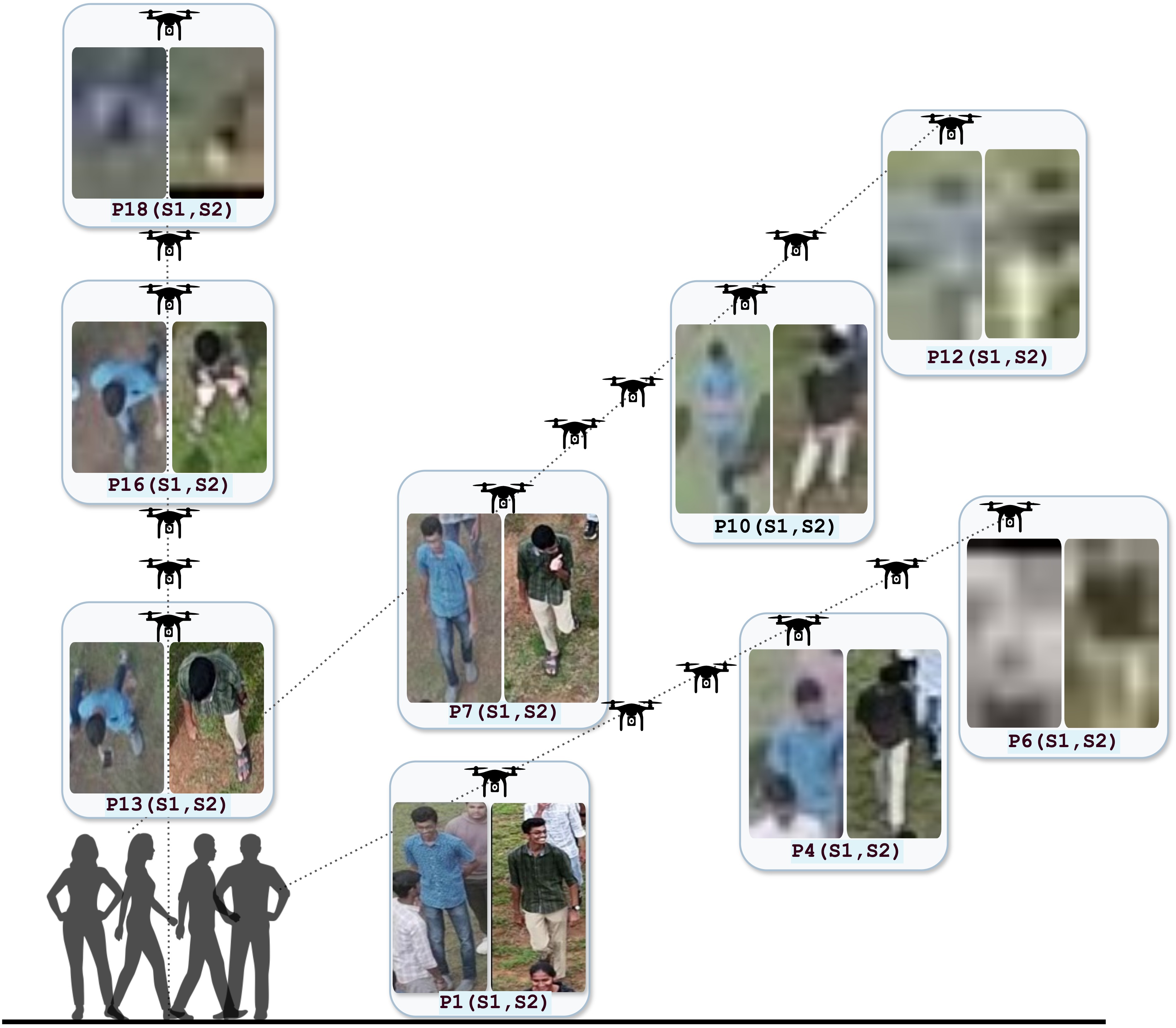}
  \caption{UAV-based outdoor capture protocol. Each subject is recorded from 18 drone viewpoints (P1–P18), spanning a wide range of altitudes, distances, and pitch angles. Recordings are repeated across two sessions (S1, S2) with varied clothing for appearance diversity.}
  \label{fig:outdoor_setup}
\end{figure}


\begin{table}[h]
  \centering
  \renewcommand{\arraystretch}{1.1}
  \setlength{\tabcolsep}{3pt} 
  \caption{Specifications of the devices used for indoor and outdoor data collection phases.}
  \label{tab:device_specs}
  \begin{tabular}{c p{1.2cm} p{0.8cm} p{0.8cm} p{1.8cm} p{1.5cm} c}
  \hline
   & \textbf{University} & \textbf{Device} & \textbf{Brand} & \textbf{Model} & \textbf{Resolution} & \textbf{FPS} \\
  \hline
  \multirow{7}{*}{\rotatebox{90}{Indoor}} 
      & UBI      & Mobile   & Apple & iPhone-14   & 2556 x 1179 & 30 \\
      & SRT      &  Mobile  & Redmi &  K50i  & 2460 x 1080 & 30 \\
      & SRM      &  Mobile  & OnePlus &    Nord CE-3& 1900 x 1400 & 30 \\
      & JNNCE    &  DSLR  & Canon &  Eos1200D  & 5184 x 3456 & 30 \\
      & MEDIPOL  &   Mobile & Apple &  iPhone-11  & 1792 x 1100 & 30 \\
      & UniLuanda   & Mobile   & Apple & iPhone-14   & 2556 x 1179 & 30 \\
      & NMDCH   &   Mobile & OnePlus & Nord CE-2   & 1900 x 1400 & 30 \\
  \hdashline
  \multirow{7}{*}{\rotatebox{90}{Outdoor}} 
      & UBI      &UAV & DJI & Phantom-4-Pro   &  3480 x 2160 & 30  \\
      & SRTMUN   & UAV & IZI &   Mini X Nano  & 5120 x 3840 & 30 \\
      & SRM      & UAV & DJI &  Mavic-3   &  4096 x 2160& 30 \\
      & JNNCE    &  UAV & DJI & Mavic-3   &  5280 x 3956& 30  \\
      & MEDIPOL  & UAV  & Piha & S155   & 2560 x 1400 &  30\\
      & UniLuanda   & UAV & DJI & Phantom-4-Pro   &  3480 x 2160 & 30  \\
      & NMDCH   & UAV & DJI & Air-S2-Fly   &  2688 x 1512 & 30  \\
  \hline
  \end{tabular}
\end{table}
\subsection{Two-Phase Collection Protocol}
\emph{DetReIDX} captures each identity through two complementary modalities:

\begin{enumerate}
    \item Indoor Capture (Ground Reference). As illustrated in Fig.~\ref{fig:indoor_setup}, each subject enrolled in this dataset undergoes i) a mugshot capture, with left profile, frontal, and right profile images; and ii) a gait video A 20-second walking sequence with turning and posture variation. Devices used at this point include DSLR and various smartphones, listed in Table~\ref{tab:device_specs}.
    \item Outdoor UAV Capture. Each subject is recorded outdoors under two sessions (S1, S2), wearing different outfits, with 18 UAV viewpoints per session. Each session captures the full range of pitch angles, altitudes, and lateral distances to introduce scale and viewpoint variance. As shown in Figure~\ref{fig:outdoor_setup} and detailed in Table~\ref{tab:drone_positions}, the drone captures include three pitch angles (30°, 60°, 90°) and six distance-altitude pairs per angle (5.8m to 120m height and 10m to 120m horizontal distance).    
\end{enumerate}

\begin{table}[h]
  \centering
  \caption{UAV capture positions and configurations. Pitch angles are defined in Session 1 and remain fixed in Session 2. Each point corresponds to a unique UAV viewpoint used in both sessions.}
  \label{tab:drone_positions}
  \begin{tabular}{cccccc}
    \hline
    \textbf{Point} & \textbf{Pitch (°)} & \multicolumn{2}{c}{\textbf{S1}} & \multicolumn{2}{c}{\textbf{S2}} \\
    \cline{3-6}
     & & \textbf{Dist. (m)} & \textbf{Height (m)} & \textbf{Dist. (m)} & \textbf{Height (m)} \\
     \hdashline
    P1  &       & 10   & 5.8   & 10   & 5.8   \\
    P2  &       & 20   & 11.5  & 20   & 11.5  \\
    P3  &       & 30   & 17.3  & 30   & 17.3  \\
    P4  & 30°   & 40   & 23.1  & 40   & 23.1  \\
    P5  &       & 80   & 40.0  & 80   & 40.0  \\
    P6  &       & 120  & 60.0  & 120  & 60.0  \\
    \hdashline
    P7  &       & 10   & 15.0  & 10   & 15.0  \\
    P8  &       & 20   & 30.0  & 20   & 30.0  \\
    P9  &       & 30   & 45.0  & 30   & 45.0  \\
    P10 & 60°   & 40   & 60.0  & 40   & 60.0  \\
    P11 &       & 80   & 75.0  & 80   & 75.0  \\
    P12 &       & 120  & 90.0  & 120  & 90.0  \\
    \hdashline
    P13 &       & 0    & 10.0  & 0    & 10.0  \\
    P14 &       & 0    & 20.0  & 0    & 20.0  \\
    P15 &       & 0    & 30.0  & 0    & 30.0  \\
    P16 & 90°   & 0    & 40.0  & 0    & 40.0  \\
    P17 &       & 0    & 80.0  & 0    & 80.0  \\
    P18 &       & 0    & 120.0 & 0    & 120.0 \\
    \hline
  \end{tabular}
\end{table}

\begin{figure*}[!t]
  \centering
  \includegraphics[width=\textwidth]{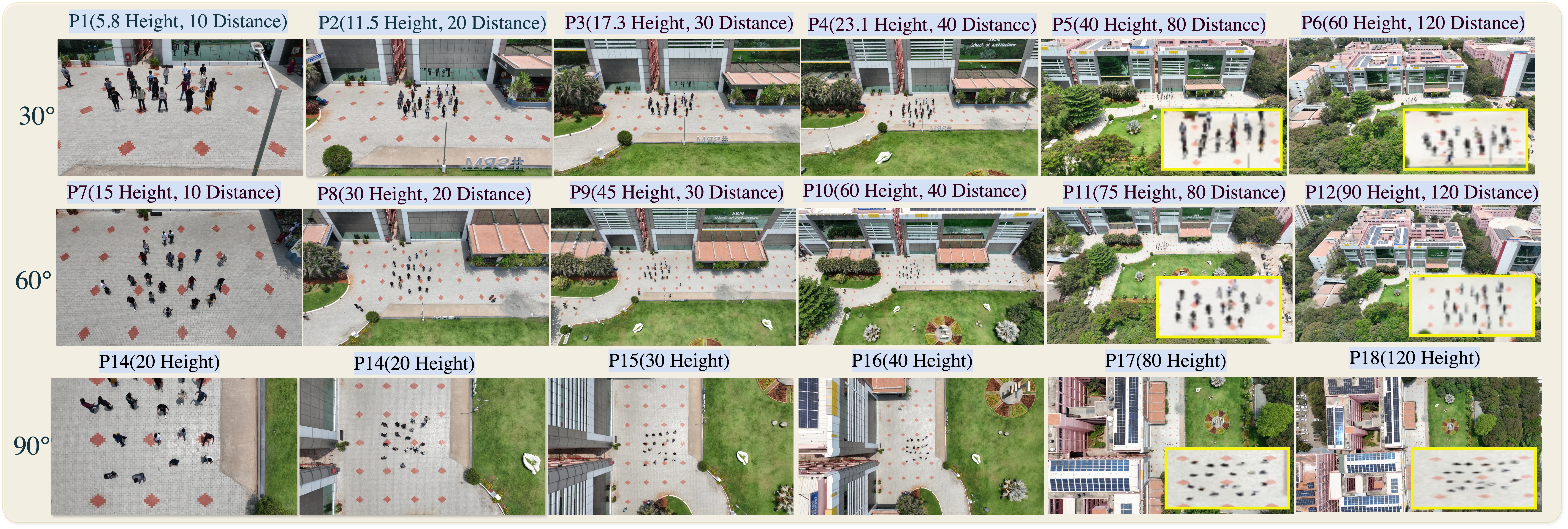}
  \caption{Actual drone-captured frames from all 18 UAV viewpoints (P1–P18), grouped by pitch angle: 30°, 60°, and 90°. Each image illustrates real-world scale variation, subject visibility, and background context. Yellow insets highlight degradation in resolution at extreme long-range positions (e.g., P6, P12, P18).}
  \label{fig:actual_drone_footage}
\end{figure*}

Subjects walk in unconstrained trajectories to simulate real-world variability. Figure~\ref{fig:actual_drone_footage} shows representative samples from all 18 viewpoints. Each video is 20+ seconds, ensuring motion, occlusion, and scale progression.

\subsection{Drone Layout and Session Design}
\begin{figure*}[h]
  \centering
  \includegraphics[width=\linewidth]{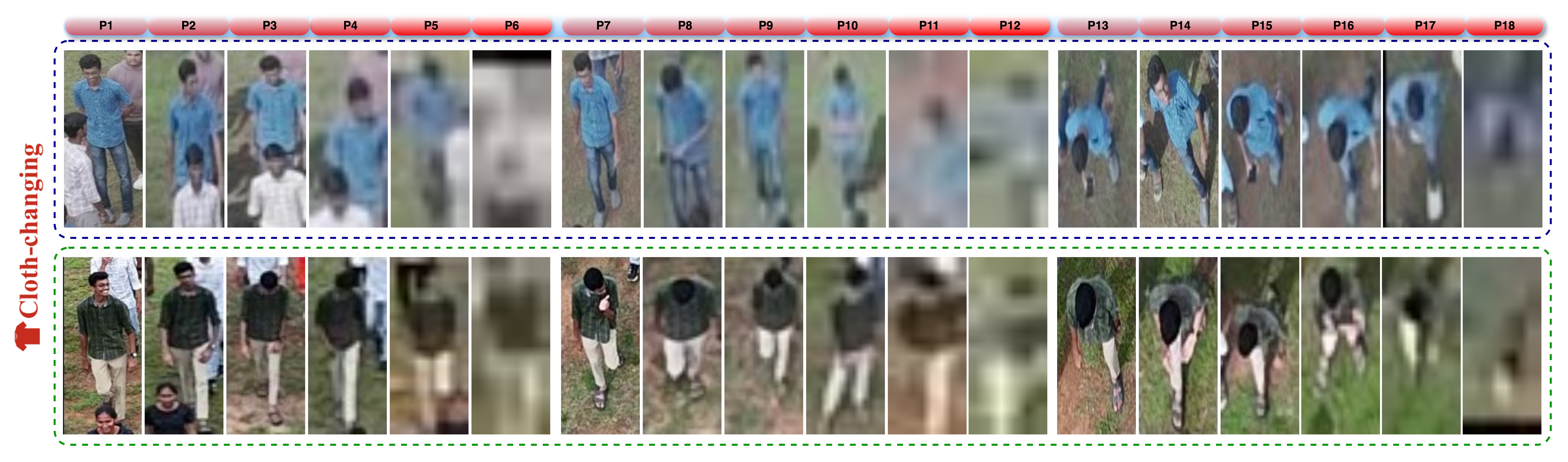}
  \caption{Example of one subject captured in 18 viewpoints (P1–P18), with clothing changes between sessions. Top row: Session 1. Bottom row: Session 2, with different attire.}
  \label{fig:DetReIDX_clothing_variation}
\end{figure*}

Each UAV flight was recorded with pitch/altitude/distance labels to support reproducible benchmark protocols. All 18 viewpoints were kept consistent across S1 and S2. This dual-session protocol aims at guaranteeing changes in appearance, particularly to guarantee that subjects wear different outfits (see Figure~\ref{fig:DetReIDX_clothing_variation}), and enable long-term ReID and clothing-insensitive search. Also, S1 and S2 were separated by at least 24 hours to ensure environmental changes (daylight, shadows, weather conditions), yielding a total of 36 drone videos per identity, divided into: i) Same-view, same-day; 
  ii) Cross-view, same-day; and iii) cross-view and cross-day, under clothing variations.

\begin{figure}[t]
  \centering
  \includegraphics[width=0.95\linewidth]{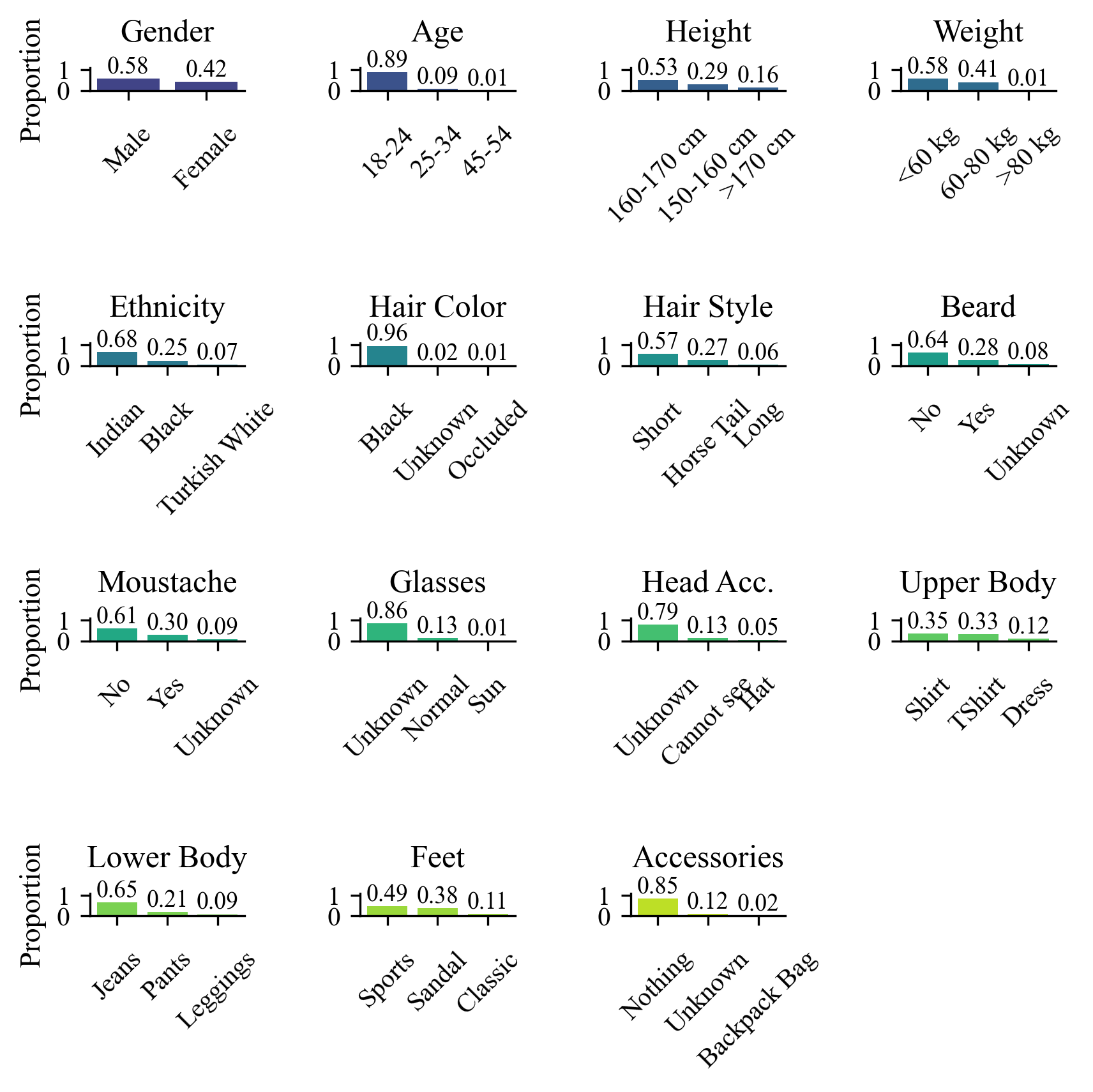}\\[1ex]
  \caption{Distributions of the soft biometric labels in \emph{DetReIDX}. The top row corresponds to the demographic distributions: the dataset is moderately male-dominated (58\% male), predominantly composed of individuals aged 18--24 (89\%), and has a high proportion of subjects in the [160, 170cm] height interval and <60kg weight ranges. Ethnic composition is skewed towards Indian (68\%) and Black (25\%) categories. The remaining rows provide different visual attributes annotated per person, including hair color, style, presence of facial hair, glasses, clothing, and accessories. Most individuals have black hair (98\%), short hairstyles (59\%), and wear normal glasses (91\%). Clothing is casual with jeans (66\%) and shirts/t-shirts being common, while accessories like bags are rare (3\%). }
  \label{fig:attribute_frequency}
\end{figure} 

\subsection{Annotation Pipeline}
All annotations were manually done by a set of volunteers, using the CVAT tool and cross-verified by peers. In total, there are 4 different kinds of annotations:

\begin{enumerate}
  \item Bounding boxes. Define each subject region-of-interest (ROI) and are annotated at fixed 10-frame intervals across all video types.
  \item Tracking IDs. Each subject is assigned a consistent PID across indoor and UAV sessions.
  \item Session metadata. Altitude, pitch, distance and scene location.
  \item Soft biometric information. 16 manual labels covering demographic, appearance, and visual cues. See Figure~\ref{fig:DetReIDX_soft_biometrics} and attribute frequency in Figure~\ref{fig:attribute_frequency}.
\end{enumerate}

Attribute completeness is benchmarked in Table~\ref{tab:attribute_comparison}, confirming that \emph{DetReIDX} offers the most detailed subject-level annotation among the aerial or cross-view related datasets.

\begin{figure}[t]
  \centering
  \includegraphics[width=\linewidth,height=6cm]{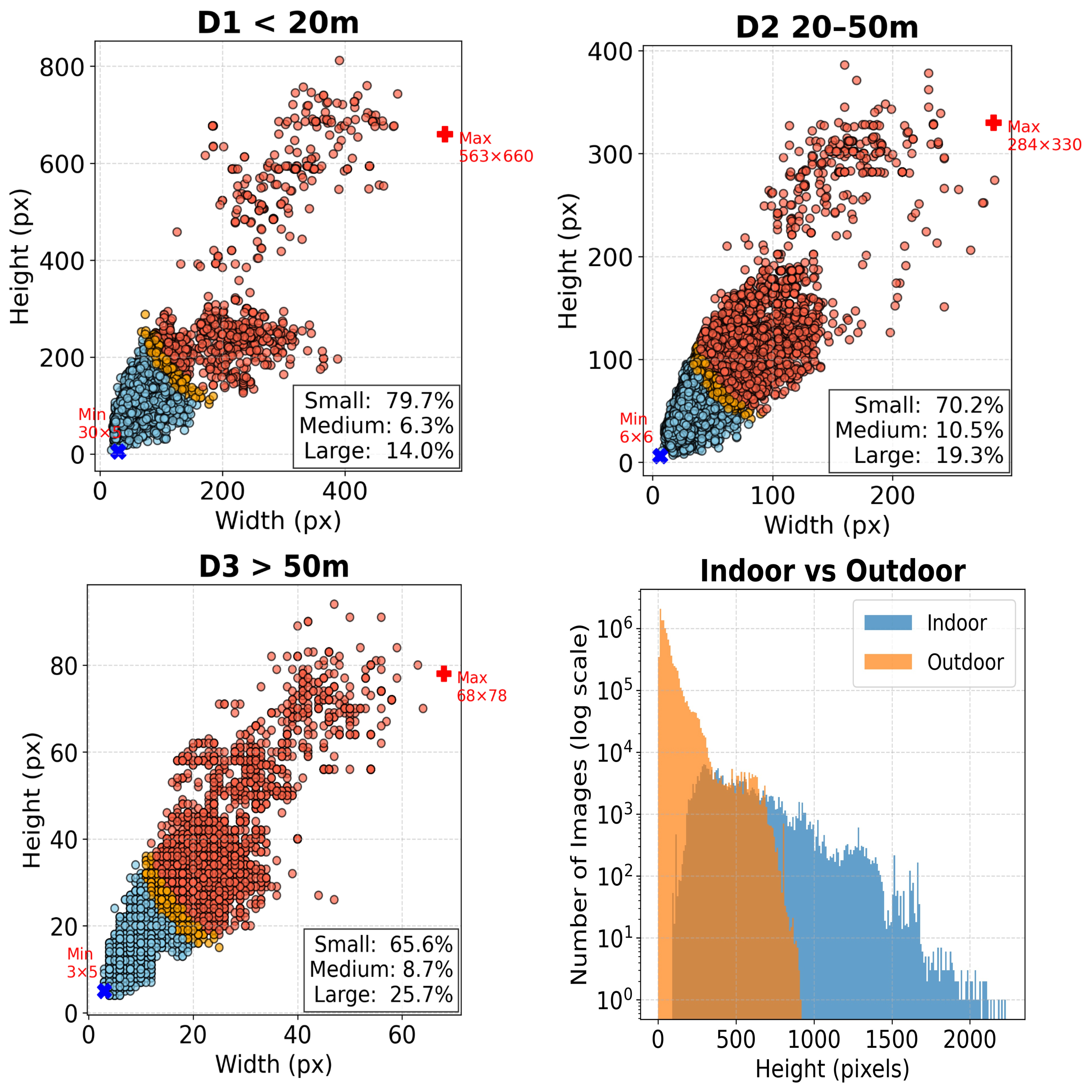}
  \caption{
    Scatter plots of ROIs height/width in three different distance bins. The bottom-right plot provides the distribution of the ROI heights (in pixels) of the indoor and outdoor data.
  }
  \label{fig:DetReIDX_resolution_variance}
\end{figure}

\begin{figure}[ht]
  \centering
  \includegraphics[width=\columnwidth]{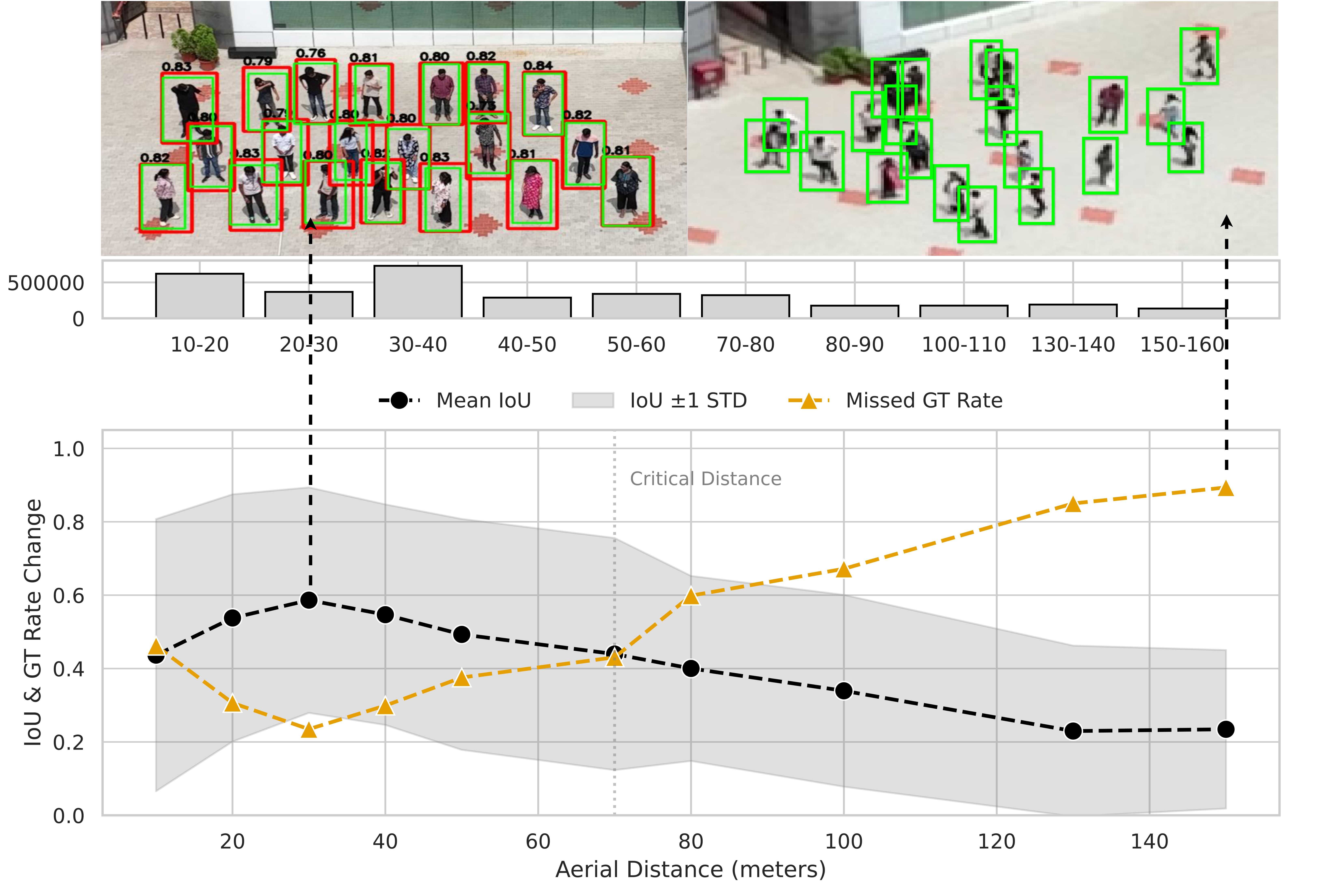}
  \caption{Effect of distance on pedestrian detection accuracy. The black curve provides the mean Intersection-over-Union (IoU) of correctly matched detections, with shaded areas representing ±1 standard deviation. The orange curve shows the proportion of missed ground truth (GT) annotations. A critical distance (70 meters) is highlighted where performance began to significantly deteriorate. The top inset visualizations illustrate example detections at \emph{close} (green box: predictions; red box: ground truth) and \emph{long} distances, corresponding to low and high GT miss rates, respectively. The bar plot above the graph indicates the number of annotations per distance bin, confirming data balance across ranges. These results  provide evidence of a substantial degradation in both detection precision and recall at long distances.
  \label{fig:distance_impact_detection}}
\end{figure}

\begin{figure}[t]
  \centering
  \includegraphics[width=\columnwidth]{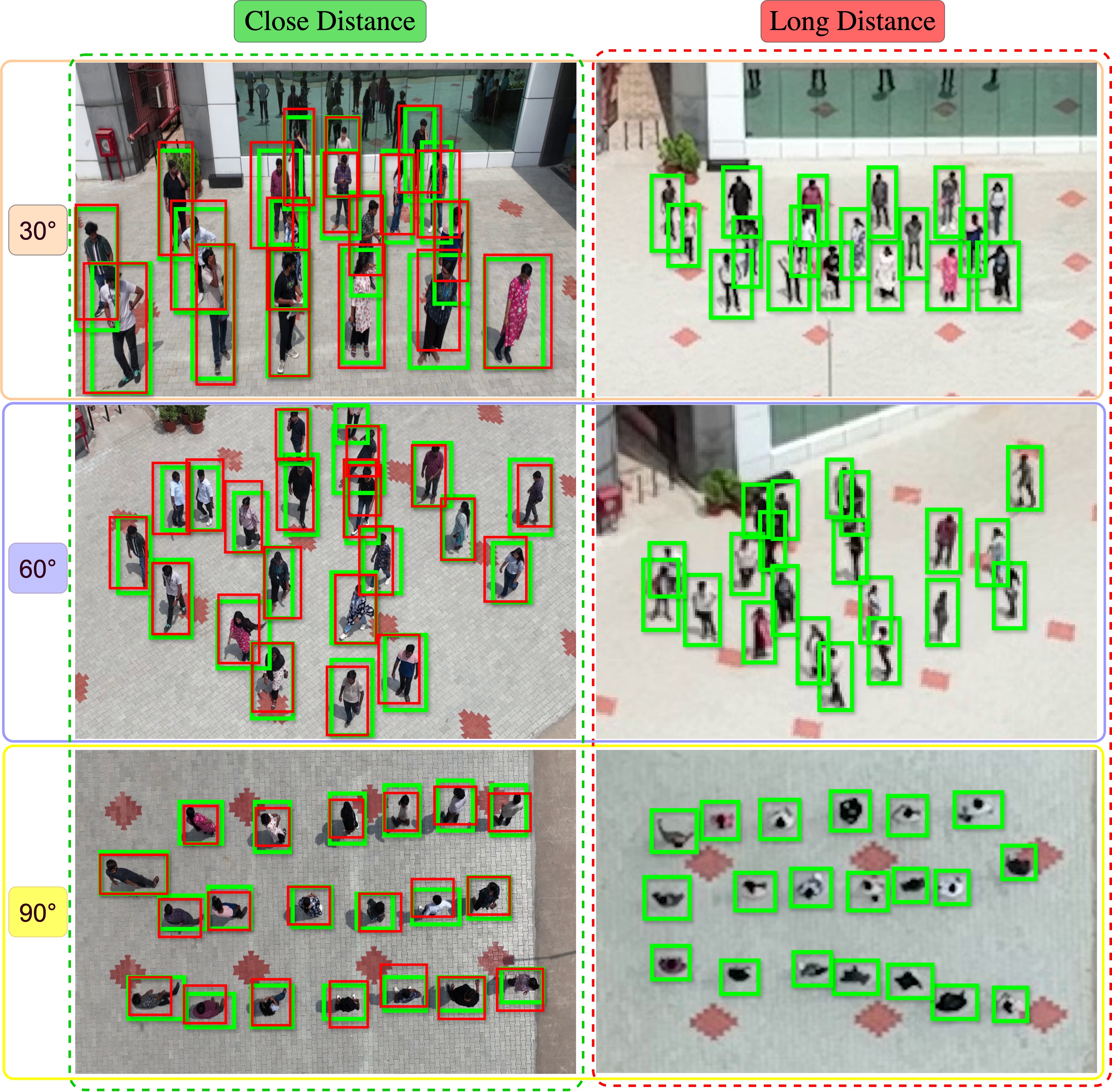}
  \caption{Qualitative analysis of pedestrian detection under varying viewpoints and distances. Rows represent different UAV pitch angles (30°, 60°, and 90°), while columns compare detections at close (left) and long ranges (right). Predicted bounding boxes from the detection model are shown in green, and ground-truth annotations are in red. As both the angle and distance increase, detection becomes more challenging due to reduced resolution, occlusion, and distortion.
  \label{fig:view_point_imapct_detection}}
\end{figure}

\begin{figure}[ht]
  \centering
  \small

  \begin{minipage}[b]{0.48\columnwidth}
    \includegraphics[width=\linewidth, height=0.40\columnwidth]{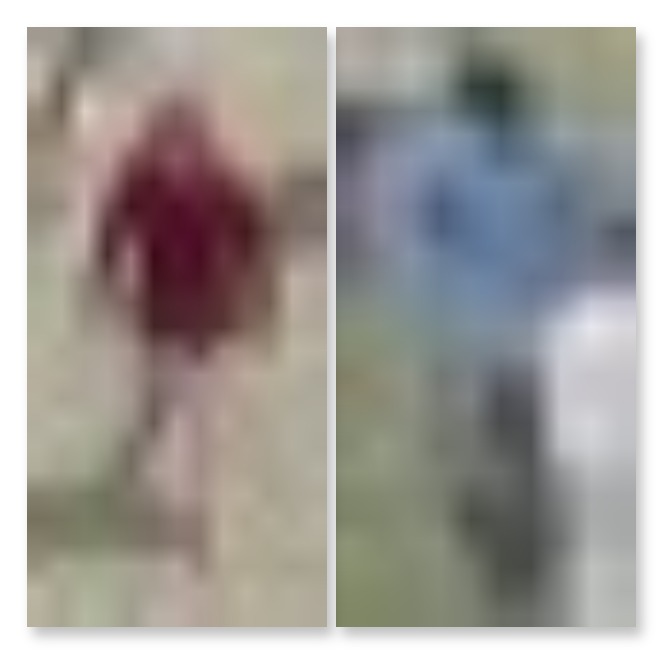}
    \centering \textbf{(a)} Low resolution
  \end{minipage}
  \hfill
  \begin{minipage}[b]{0.48\columnwidth}
    \includegraphics[width=\linewidth, height=0.40\columnwidth]{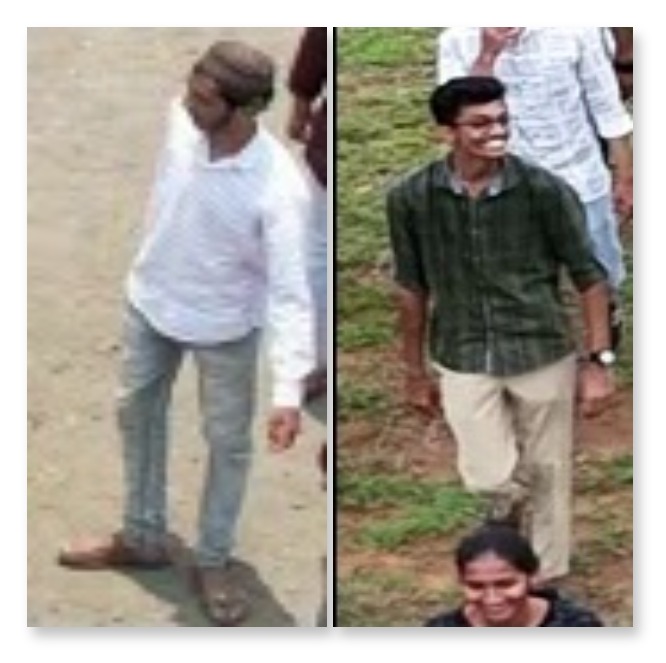}
    \centering \textbf{(b)} Clothing variation
  \end{minipage}

  \vspace{4pt}

  \begin{minipage}[b]{0.48\columnwidth}
    \includegraphics[width=\linewidth, height=0.40\columnwidth]{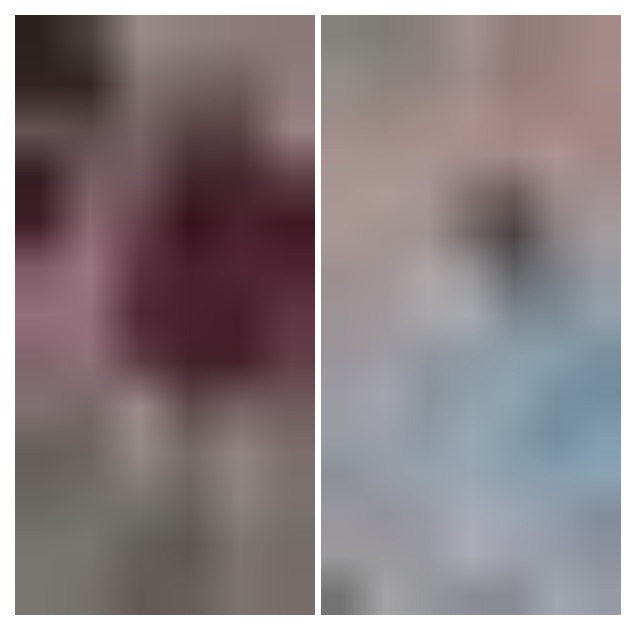}
    \centering \textbf{(c)} Long-range
  \end{minipage}
  \hfill
  \begin{minipage}[b]{0.48\columnwidth}
    \includegraphics[width=\linewidth, height=0.40\columnwidth]{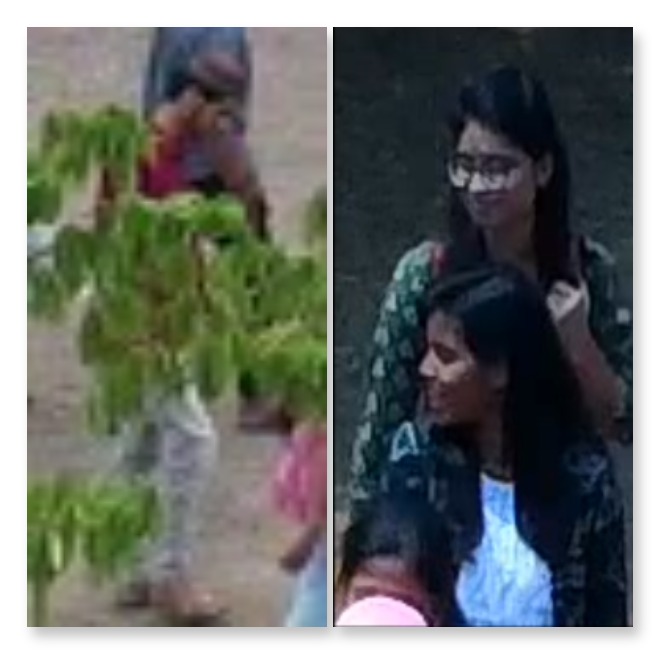}
    \centering \textbf{(d)} Occlusion
  \end{minipage}

  \vspace{4pt}

  \begin{minipage}[b]{0.48\columnwidth}
    \includegraphics[width=\linewidth, height=0.40\columnwidth]{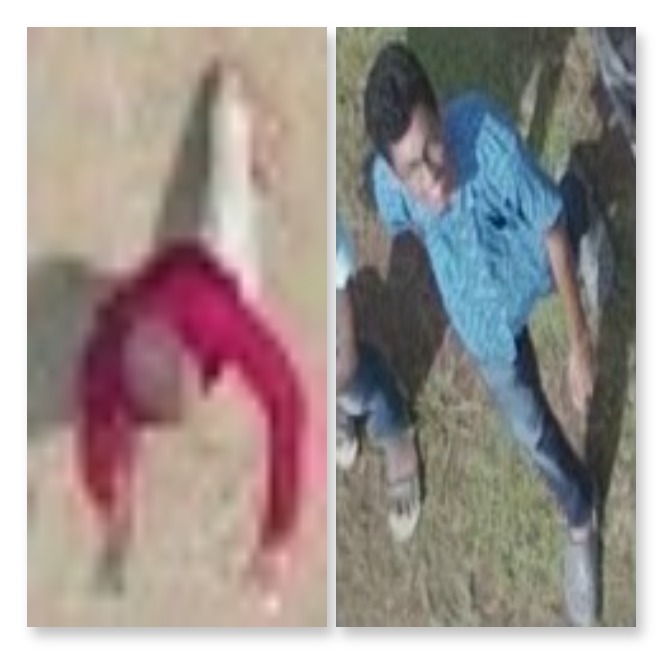}
    \centering \textbf{(e)} Top-down view
  \end{minipage}
  \hfill
  \begin{minipage}[b]{0.48\columnwidth}
    \includegraphics[width=\linewidth, height=0.40\columnwidth]{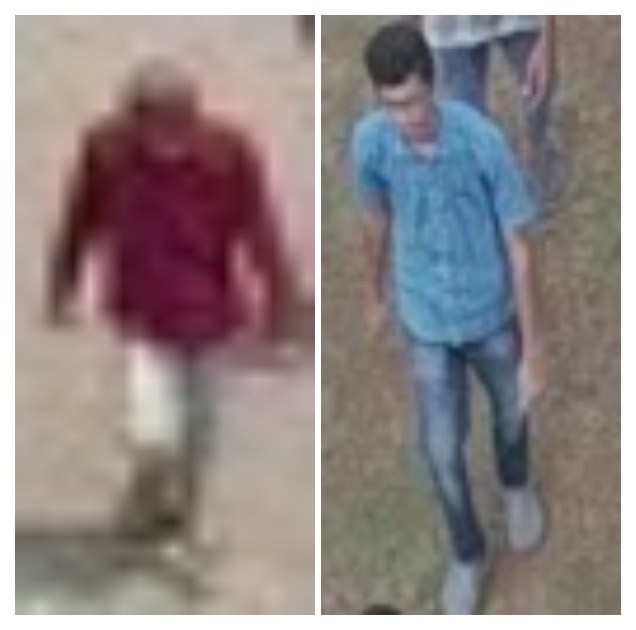}
    \centering \textbf{(f)} Pose variation
  \end{minipage}

  \vspace{4pt}

  \begin{minipage}[b]{0.48\columnwidth}
    \includegraphics[width=\linewidth, height=0.40\columnwidth]{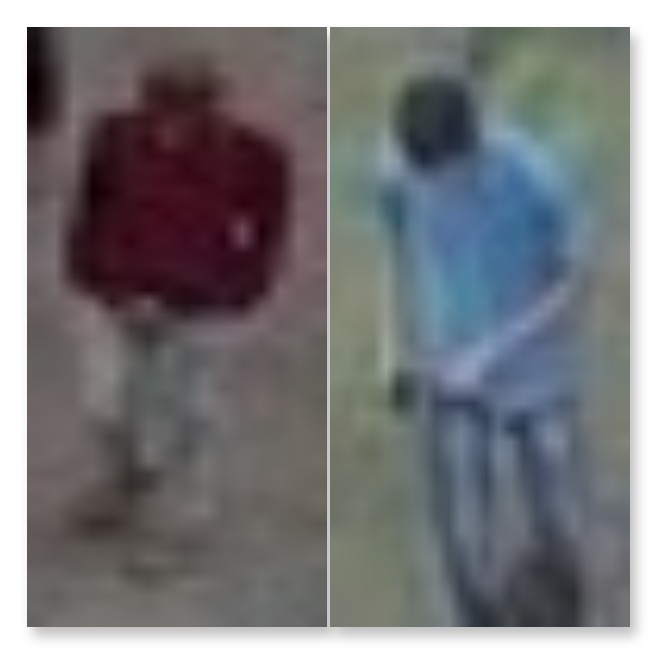}
    \centering \textbf{(g)} Motion blur
  \end{minipage}
  \hfill
  \begin{minipage}[b]{0.48\columnwidth}
    \includegraphics[width=\linewidth, height=0.40\columnwidth]{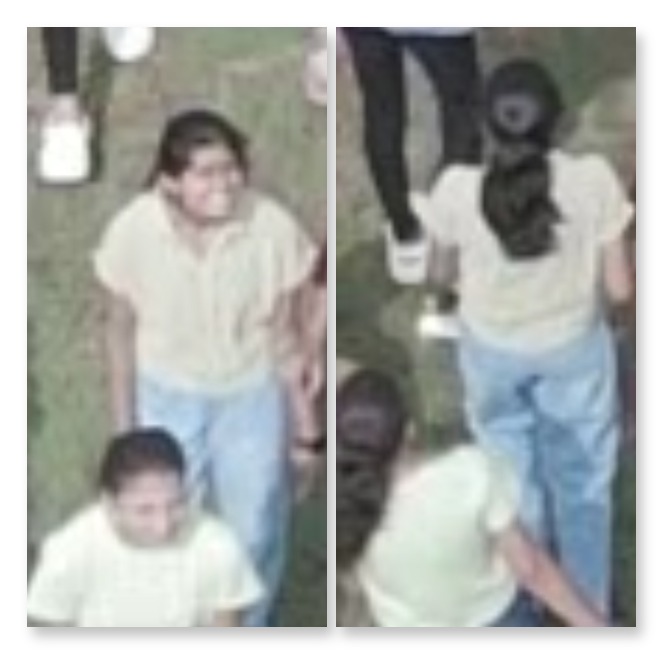}
    \centering \textbf{(h)} View perspective
  \end{minipage}

  \caption{Challenging conditions in person identification from UAV footage: (a) low resolution, (b) clothing variation, (c) long-range observations, (d) occlusion, (e) top-down viewpoints, (f) pose variation, and (g) motion blur.}
  \label{fig:DetReIDX_challenges}
\end{figure}

\subsection{Viewpoint and Resolution Diversity}
As shown in Figure~\ref{fig:DetReIDX_resolution_variance}, pedestrian scale varies drastically across UAV positions. Indoor captures often exceed 1000px bounding box height, while aerial views in P18 (90°, 120m) provide ROIs smaller than 10px tall, approaching scale-invariant detection limits.

Figure~\ref{fig:view_point_imapct_detection} and Figure~\ref{fig:DetReIDX_challenges} illustrate how UAV angle and altitude lead to occlusion, distortion, and viewpoint-specific degradation. \emph{DetReIDX} captures this with pixel-level granularity, enabling fine-grained robustness evaluation.

\begin{table}[h]
  \centering
  \caption{\emph{DetReIDX} Outdoor Dataset Statistics}
  \begin{tabular}{lrrrl}
  \hline
  \textbf{Split} & \textbf{\#Videos} & \textbf{\#Images} & \textbf{\#Annotations} & \textbf{Formats} \\
  \hline
  Train       & 120 & 131,580 & 5,095,539 & YOLO, COCO \\
  Validation  & 56  & 63,591  & 2,483,836 & YOLO, COCO \\
  Test        & 109 & 108,252 & 4,217,824 & YOLO, COCO \\
  \hline
  Total       & 285 & 303,423 & 11,797,199 &  \\
  \hline
  \end{tabular}
  \label{tab:DetReIDX_split}
\end{table}

\begin{table}[t]
  \centering
  \caption{Statistics of the \emph{DetReIDX} ReID data splits, for the Aerial $\rightarrow$ Aerial, Aerial $\rightarrow$ Ground  and Ground $\rightarrow$ Ground settings.}
  \label{tab:DetReIDX_reid}
  \begin{tabular}{lrrr}
    \toprule
    \textbf{Split / Test Case} & \textbf{\#Query} & \textbf{\#Gallery} & \textbf{Total Images} \\
    \midrule
    Train (Indoor + Outdoor) & --        & --        & 289,392 \\
    \hdashline
    Aerial $\rightarrow$ Aerial        & 52,926    & 52,552    & 105,478 \\
    Aerial $\rightarrow$ Ground         & 106,927   & 7,959     & 114,886 \\
    Ground $\rightarrow$ Aerial         & 7,959     & 106,927   & 114,886 \\
    \bottomrule
  \end{tabular}
\end{table}

\subsection{Data Splits and Formats}
\emph{DetReIDX} annotations are released in YOLO and COCO formats. ReID queries and galleries are organized for aerial-to-aerial (A→A), aerial-to-ground (A→G), and ground-to-aerial (G→A) matching settings (Table~\ref{tab:DetReIDX_reid_stats}). Detection splits (Table~\ref{tab:DetReIDX_split}) follow scene- and viewpoint-aware partitioning, with no video overlap between train and test.

\subsection{\emph{DetReIDX} Uniqueness}

As stated above, DetReIDX was designed to fill the most important key blind spots in current pedestrian recognition research, enabling: a) cross-domain ReID,  by matching UAV views to high-resolution indoor references (A→G); b) clothing-invariant search, with clothing changes \emph{within-subject} between the different sessions; c) long-range detection, with UAV-to-subject distances up to 120m (Figure~\ref{fig:distance_impact_detection}); and d) extreme low-resolution and severe occlusions, with pedestrian ROIs as small as 8×8 pixels (Figure~\ref{fig:DetReIDX_challenges}a).

Table~\ref{tab:dataset_comparison} presents a side-by-side breakdown of \emph{DetReIDX} versus the leading ground-ground (e.g., Market-1501~\cite{zheng2015scalable}, Duke~\cite{wu2018exploit}), aerial-aerial (e.g., UAV-Human~\cite{li2021uav}, P-DESTRE~\cite{kumar2020p}), and aerial-ground (e.g., AG-ReID.v2~\cite{nguyen2024ag}, G2APS~\cite{wang2025secap}) datasets.

\subsection{Ethical Considerations}
All participants gave their informed consent in writing. Data was anonymized where necessary. \emph{DetReIDX} include facial detail and is released under a non-commercial research license for academic use. UAV flights were approved by institutional review boards and followed any existing local regulations.

\section{Experiments and Results}
\label{sec:experiments}

As a primary benchmark of the dataset, we conducted extensive experiments to assess performance of state-of-the-art (SOTA) models in pedestrian detection and re-identification (ReID) tasks. Each evaluation setting was designed to evaluate model robustness across realistic surveillance variables: altitude, angle, range, resolution, and cross-domain identity transfer.

\subsection{Pedestrian Detection}

Being at the basis of the ReID pipeline, pedestrian detection actual sustains the whole process, as any failures will compromise any subsequent phase. Also, as it is typically the earliest processing phase, it is the one that first should handle the dynamics of the environments. For this case, only he outdoor subset of \emph{DetReIDX} was considered challenging enough, including 285 UAV video sequences. We used a 70-20-10 split for training, validation, and testing, with absolutely no overlap across splits. 

As baselines, we selected three pedestrian detectors that we consider to represent the SOTA: i) YOLOv8~\cite{roboflow2023yolov8}: an anchor-free one-stage detector with decoupled heads; ii) DDOD~\cite{chen2021disentangle}, a disentangled dense object detector addressing label assignment and scale bias; and iii) Grid-RCNN~\cite{lu2019grid}: a region-based detector using pixel-level grid point prediction. Each model was trained from scratch on the \emph{DetReIDX} training set, and evaluated using the AP@50 (IoU) performance metric.

Two main factors were identified as the most obvious covariates for human detection performance: viewpoint (perspective) and distance (scale). Then, being particularly important to understand the generalization capabilities of the different methods, our experiments mainly assume the \emph{interpolation} and \emph{extrapolation}, depending whether the test viewpoints/distances are (aren't) enclosed in the corresponding learning intervals. 

At first, as baseline performance, all pitch angles (30\degree, 60\degree, 90\degree) and distances were used for training and test purposes. Then, to perceive the viewpoint generalization performance, two modes were tested: i) Interpolation (30\degree, 90\degree $\rightarrow$ 60\degree), with models trained on extreme angles and tested on the mid-views; and the more challenging ii) Extrapolation (30\degree, 60\degree $\rightarrow$ 90\degree):, where tests are done on unseen extreme views. Regarding distance generalization, we quantized the acquisition distances into three bins: D1: $<$20m (short-range); D2: 20--50m (mid-range); and D3: $>$50m (long-range). Next, in a similar way to viewpoint, these splits were used to train/test across distance bins and evaluate the robustness of SOTA models across scale.

\begin{table*}[t]
  \centering
  \caption{AP50 of YOLOv8, DDOD, and Grid-RCNN on the \emph{DetReIDX} dataset across aerial viewpoint and distance range shifts. Scores are reported as absolute AP50 followed by percentage change from the baseline (↑: gain, ↓: drop).}
  \label{tab:DetReIDX_final_d123}
  \setlength{\tabcolsep}{5pt}
  \renewcommand{\arraystretch}{1.1}
  \begin{tabular}{lcc|c|c|c}
    \toprule
    \textbf{Experiment} & \textbf{Train Set} & \textbf{Test Set} 
    & \textbf{YOLOv8} & \textbf{DDOD} & \textbf{Grid-RCNN} \\
    \midrule

    Baseline (All Conditions) & ALL & ALL &
    0.734 & 0.608 & 0.620 \\
    \hdashline
    Interpolation  & 30°, 90° & 60° &
    \textcolor{red}{0.669 (↓8.90\%)} &
    \textcolor{red}{0.564 (↓7.20\%)} &
    \textcolor{red}{0.514 (↓17.1\%)} \\

    Extrapolation  & 30°, 60° & 90° &
    \textcolor{red}{0.503 (↓31.5\%)} &
    \textcolor{red}{0.474 (↓22.0\%)} &
    \textcolor{red}{0.403 (↓35.0\%)} \\
    \hdashline

     (D1 → D1) &  & D1 &
    \textcolor{green!50!black}{0.914 (↑24.5\%)} &
    \textcolor{green!50!black}{0.857 (↑40.9\%)} &
    \textcolor{green!50!black}{0.839 (↑35.3\%)} \\

    (D1 → D2) & D1 & D2 &
    \textcolor{green!50!black}{0.793 (↑8.00\%)} &
    \textcolor{red}{0.380 (↓37.5\%)} &
    \textcolor{red}{0.428 (↓30.9\%)} \\

    (D1 → D3) &  & D3 &
    \textcolor{red}{0.137 (↓81.3\%)} &
    \textcolor{red}{0.008 (↓98.7\%)} &
    \textcolor{red}{0.009 (↓98.5\%)} \\
    \hdashline
    (D2 → D1) &  & D1 &
    \textcolor{red}{0.694 (↓5.50\%)} &
    \textcolor{red}{0.582 (↓4.30\%)} &
    \textcolor{green!50!black}{0.668 (↑7.70\%)} \\

     (D2 → D2) & D2 & D2 &
    \textcolor{green!50!black}{0.890 (↑21.2\%)} &
    \textcolor{green!50!black}{0.776 (↑27.6\%)} &
    \textcolor{green!50!black}{0.770 (↑24.2\%)} \\

    (D2 → D3) &  & D3 &
    \textcolor{red}{0.315 (↓57.1\%)} &
    \textcolor{red}{0.111 (↓81.8\%)} &
    \textcolor{red}{0.150 (↓75.8\%)} \\
    \hdashline
    (D3 → D1) &  & D1 &
    \textcolor{red}{0.015 (↓97.9\%)} &
    \textcolor{red}{0.004 (↓99.3\%)} &
    \textcolor{red}{0.002 (↓99.7\%)} \\

    (D3 → D2) & D3 & D2 &
    \textcolor{red}{0.411 (↓44.0\%)} &
    \textcolor{red}{0.274 (↓54.9\%)} &
    \textcolor{red}{0.261 (↓57.9\%)} \\

     (D3 → D3) &  & D3 &
    \textcolor{red}{0.581 (↓20.8\%)} &
    \textcolor{red}{0.408 (↓32.9\%)} &
    \textcolor{red}{0.280 (↓54.8\%)} \\

    \bottomrule
  \end{tabular}
\end{table*}

Table~\ref{tab:DetReIDX_final_d123} summarises the observed AP@50 values. As key observations, we highlight several notable cases: a) long-range collapse (D1$\rightarrow$D3): YOLOv8 drops from 91.4\% (D1$\rightarrow$D1) to 13.7\% (D1$\rightarrow$D3), and DDOD/GR-CNN degrade by 90\%+. Detection fails entirely at >50m due to sub-10 pixel targets; b) Viewpoint Failure (Extrapolation): All models perform significantly worse on unseen 90\degree{} top-down views, highlighting angular overfitting; and c) Reverse Transfer Limits: D3$\rightarrow$D1 performance is near zero, indicating that models trained only on long-range views are not able to learn transferable pedestrian features. Figures~\ref{fig:view_point_imapct_detection} and~\ref{fig:distance_impact_detection} illustrate how performance deteriorates with increasing pitch and distance due to object scale collapse, blur, and top-down foreshortening.

\subsection{Pedestrian Re-Identification}
\label{sec:reid}

The \emph{DetReIDX} benchmark introduces a high-fidelity ReID testbed simulating real-world aerial-ground surveillance, where most conventional ReID assumptions break down. It contains 509 unique identities recorded indoors, of which 334 (65.6\%) are re-observed in outdoor UAV scenes. Each subject appears in at least two recording sessions with different clothing and variable lighting, enabling cross-session, cross-domain ReID evaluation.

A 70\%-30\% PID-disjoint train-test split is used, assigning 267 identities (289{,}392 images) to training and 67 identities (114{,}886 images) to testing. Each test identity is captured across 36 UAV video sequences (two sessions × 18 aerial viewpoints) and one controlled indoor gait video, enabling high-variance retrieval under extreme appearance, angle, and resolution variation.

We define three canonical test scenarios:

\begin{itemize}
    \item Aerial$\rightarrow$Aerial (A2A): Queries are UAV sequences from Session 1; gallery samples from Session 2. This isolates cross-session variation within the aerial domain.
    \item Aerial$\rightarrow$Ground (A2G): UAV-based queries are matched against high-quality indoor references. This tests cross-domain generalization from in-the-wild to controlled settings.
    \item Ground$\rightarrow$Aerial (G2A): Indoor queries are matched against UAV galleries. This tests downward domain transfer.
\end{itemize}

The statistics of each scenario are listed in Table~\ref{tab:DetReIDX_reid}, and all of them were evaluated using the same metrics: Rank-1, Rank-5, Rank-10, and mean Average Precision (mAP). 

\begin{table}[t]
  \centering
  \caption{\emph{DetReIDX} ReID split statistics.}
  \label{tab:DetReIDX_reid_stats}
  \begin{tabular}{lrrr}
    \toprule
    \textbf{Scenario} & \textbf{\#Query} & \textbf{\#Gallery} & \textbf{Total Images} \\
    \midrule
    Train (Indoor + UAV) & --        & --        & 289,392 \\
    \hdashline
    A2A (UAV$\rightarrow$UAV) & 52,926 & 52,552 & 105,478 \\
    A2G (UAV$\rightarrow$Indoor) & 106,927 & 7,959 & 114,886 \\
    G2A (Indoor$\rightarrow$UAV) & 7,959 & 106,927 & 114,886 \\
    \bottomrule
  \end{tabular}
\end{table}

Again, as baselines, we selected three recent ReID methods considered to represent the SOTA: a) PersonViT~\cite{hu2025personvit}: a transformer-based model trained on large-scale ReID datasets using global attention across spatial features; b) SeCap~\cite{wang2025secap}, an aerial-aware model using spatially enhanced capsule networks to align features across drone-ground domains; and c) 
    CLIP-ReID~\cite{li2023clip}: a vision-language pretrained CLIP model, adapted here for image-only ReID using prompt-based fine-tuning.

As shown in Table~\ref{tab:reid_results}, all models perform poorly across \emph{DetReIDX} test conditions. Despite the relatively good performance on the existing ground-level datasets, no  model was observed  to generalize to \emph{DetReIDX}'s real-world constraints.

\begin{table}[t]
  \centering
  \caption{Overall ReID performance observed on the \emph{DetReIDX} dataset.}
  \label{tab:reid_results}
  \footnotesize
  \begin{tabular}{llrrrr}
    \toprule
    \textbf{Model} & \textbf{Scenario} & \textbf{mAP (\%)} & \textbf{R1 (\%)} & \textbf{R5 (\%)} & \textbf{R10 (\%)} \\
    \midrule
    \multirow{3}{*}{PersonViT}
      & A2A & 9.9 & 8.8 & 14.4 & 17.6 \\
      & A2G & 22.3 & 19.6 & 24.8 & 27.6 \\
      & G2A & 23.3 & 51.9 & 59.4 & 63.0 \\
    \hdashline
    \multirow{3}{*}{SeCap}
      & A2A & 11.2 & 8.2 & 13.0 & 16.2 \\
      & A2G & 20.5 & 18.1 & 21.5 & 23.4 \\
      & G2A & 21.2 & 50.9 & 57.7 & 60.7 \\
    \hdashline
    \multirow{3}{*}{CLIP-ReID}
      & A2A & 9.5 & 8.9 & 12.8 & 15.3 \\
      & A2G & 22.0 & 19.7 & 24.0 & 26.2 \\
      & G2A & 20.8 & 58.1 & 63.1 & 65.2 \\
    \bottomrule
  \end{tabular}
\end{table}

\subsubsection{Qualitative Analysis}
\begin{figure*}[t]
  \centering
  \includegraphics[width=\textwidth]{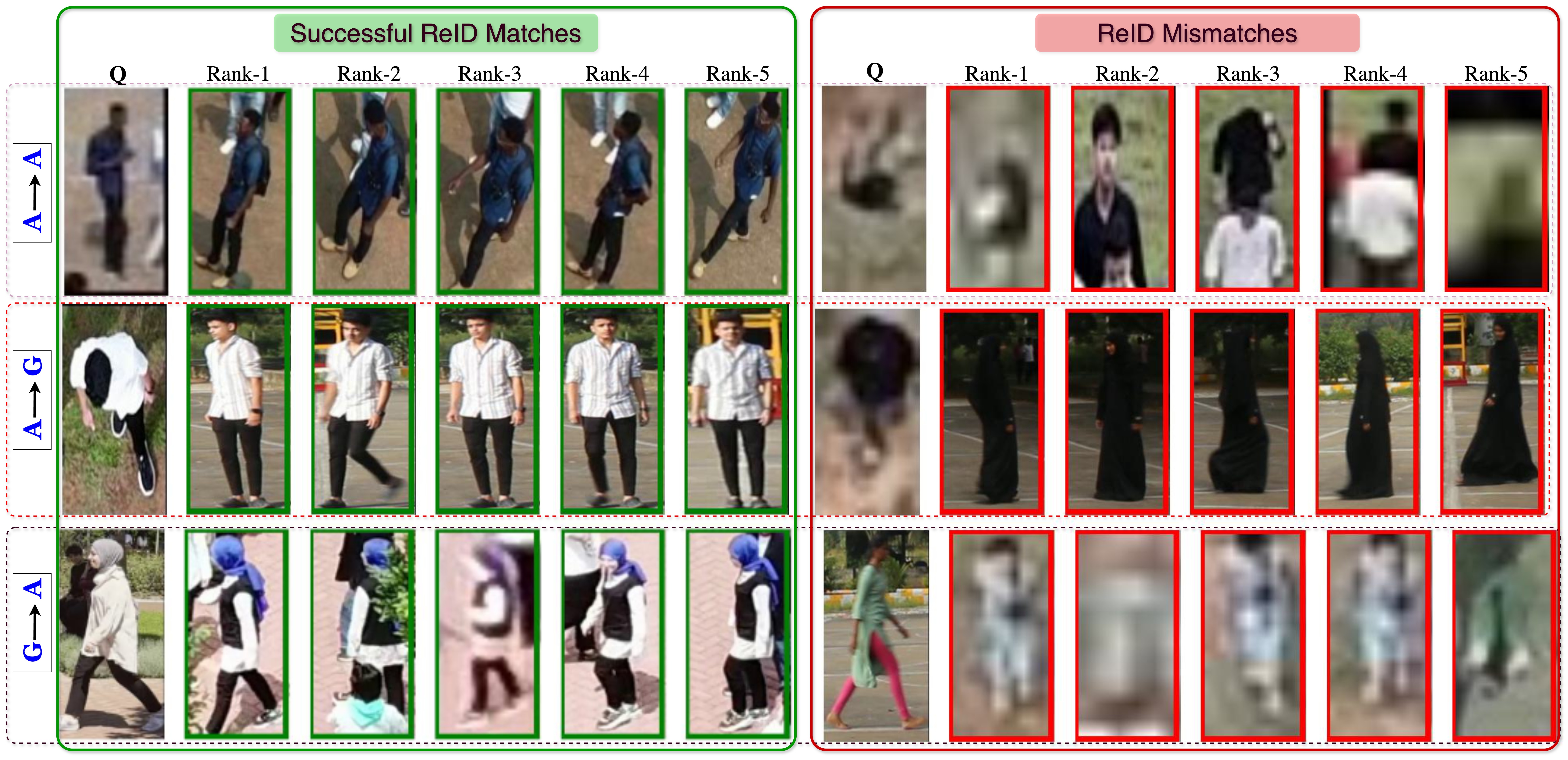}
  \caption{Qualitative evaluation of Person-ViT ReID model on \emph{DetReIDX} dataset. The left panel (green) illustrates successful retrieval cases where UAV-based query images ("Q") yield correct matches among top-5 retrieved identities (Rank-1 to Rank-5). The right panel (red) shows failure cases highlighting typical conditions challenging ReID performance, including severe aerial-to-ground (A→G), aerial-to-aerial (A→A), and ground-to-aerial (G→A) viewpoint changes, extreme long-range resolution loss, significant appearance variations due to clothing changes across recording sessions, and environmental factors such as motion blur and occlusion. These results underline the limitations of current state-of-the-art models in real-world UAV surveillance scenarios, as explicitly addressed by the \emph{DetReIDX} dataset.}
  \label{fig:reid_rank}
\end{figure*}

 Figure~\ref{fig:reid_rank} provides some remarkable examples, that were considered to represent the typical failure/success cases. In general, successful retrievals (left) tend to occur under the following conditions: consistent clothing, relatively low altitudes, and low variable silhouette profiles. On the other way, the right side of the figure illustrates the typical failure cases, mostly due to severe occlusions, low resolution, extreme pitch, and clothing changes.

\subsubsection{Impact of UAV Altitude on Retrieval}
\begin{figure*}[t]
  \centering
  \includegraphics[width=\textwidth]{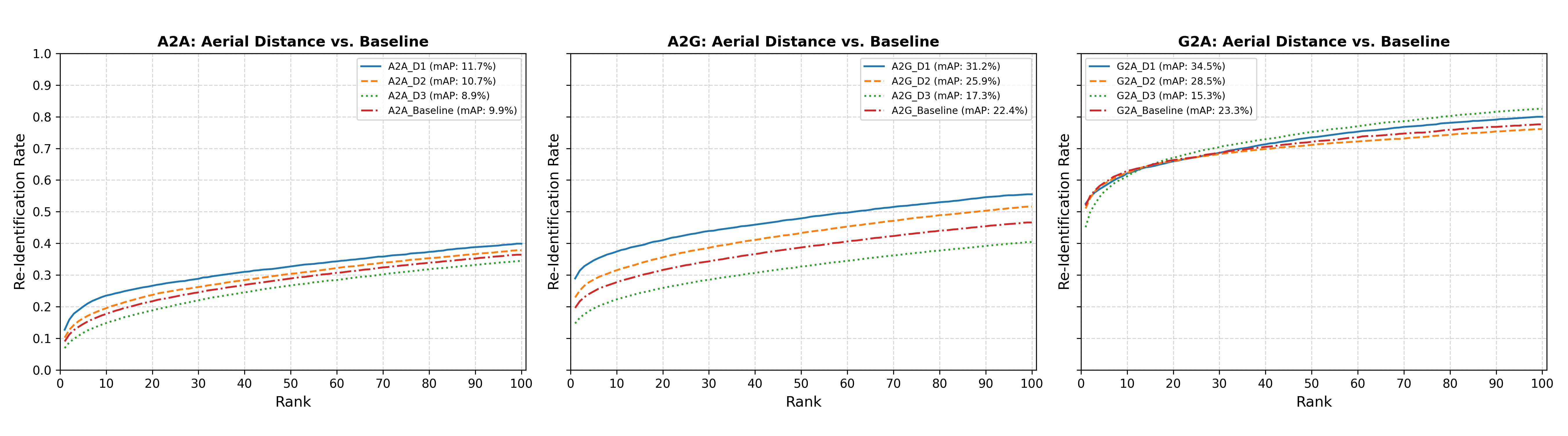}
  \caption{Cumulative Match Characteristic (CMC) curves showing the impact of aerial distances on ReID performance using the Person-ViT model, evaluated across three domain transfer scenarios provided by the \emph{DetReIDX} dataset: A2A, A2G, and G2A. Each scenario compares retrieval performance at different aerial distance intervals: close-range (D1: $<$20m), mid-range (D2: 20–50m), and long-range (D3: $>$50m) against an all-distance baseline. Results highlight significant degradation in ReID accuracy with increasing aerial distance due to factors such as severe resolution loss, viewpoint distortion, and reduced discriminative appearance features. Mean Average Precision (mAP) scores provided in the legends quantify performance drops, emphasizing long-range recognition challenges specifically targeted by \emph{DetReIDX}.}
  \label{fig:distance_vs_baseline}
\end{figure*}

To isolate aerial viewpoint effects, we quantized the queries by drone distance (D1: low, D2: medium, D3: high altitude). Table~\ref{tab:reid_distance_impact} and Figure~\ref{fig:distance_vs_baseline} reveal a consistent performance collapse with altitude across all tasks. For instance, in A2G, mAP drops from 31.2\% (D1) to 17.3\% (D3).

\begin{table}[t]
  \centering
  \caption{ReID performance by UAV distance (D1–D3).}
  \label{tab:reid_distance_impact}
  \footnotesize
  \begin{tabular}{cccccc}
    \toprule
    \textbf{Scenario} & \textbf{Distance} & \textbf{mAP} (\%) & \textbf{R1} (\%)& \textbf{R5} (\%)& \textbf{R10} (\%)\\
    \midrule
    \multirow{3}{*}{A2A} 
      & D1 & 11.7 & 12.7 & 20.0 & 23.5 \\
      & D2 & 10.7 & 10.2 & 16.3 & 19.5 \\
      & D3 & 8.9 & 6.9 & 11.7 & 14.8 \\
    \hdashline
    \multirow{3}{*}{A2G} 
      & D1 & 31.2 & 28.9 & 34.6 & 37.4 \\
      & D2 & 25.9 & 22.9 & 28.5 & 31.5 \\
      & D3 & 17.3 & 14.7 & 19.4 & 22.3 \\
    \hdashline
    \multirow{3}{*}{G2A} 
      & D1 & 34.5 & 52.5 & 58.0 & 62.1 \\
      & D2 & 28.5 & 51.0 & 58.8 & 62.3 \\
      & D3 & 15.3 & 45.1 & 56.3 & 61.2 \\
    \bottomrule
  \end{tabular}
\end{table}

\subsubsection{Failure cases and Futher Research}

According to our experiments, \emph{DetReIDX} exposes critical blind spots in the existing SOTA Re-ID models. In particular, we emphasize: a) the viewpoint dependency: Overhead UAV angles eliminate body and gait structure; b) clothing reliance: Appearance drift invalidates color- or texture-based cues; c) resolution limits: Long-range views reduce pedestrians to <20px silhouettes; and d) domain disjointness: with indoor and UAV domains yielding notorious feature mismatch.

This way, to improve the results in the \emph{DetReIDX}, any forthcoming generation of models should keep as priorities:

\begin{itemize}
    \item Learn viewpoint-agnostic representations robust to pitch and elevation. The subjects appearance varies dramatically with respect to  pitch angles, in particular. It is up to the models to identify and register specific correspondences between data acquired from different perspectives. 
    \item Achieve resolution invariance. The current generation of methods tends to rely on minutiae information to obtain appropriate feature representations. However, for very small resolutions (e.g., <15px targets) such kind of information isn't discernible.  
    \item Focus on soft biometrics or geometry-aware features over appearance-based information, which is much sensitive to daylight and perspective.   
    \item Obtain cross-domain registration between UAV and controlled views data, which is particularly important to match data acquired from very different sensors, or even different light spectra. 
\end{itemize}

\section{Conclusions}
\label{sec:conclusion}

Due to safety/security concern in modern societies, person ReID from surveillance footage has been establishing as technology of particular interest. However, we observed that SOTA methods catastrophically fail when facing actual \emph{real-world} conditions, such as extreme pitch angles, long-range scale distortions, appearance drifts, and tiny resolution.

This observation was the primary motivation for the development of the \emph{DetReIDX} dataset, which purposely integrates such variability factors by design. Spanning 5.8--120m altitudes, 18 aerial viewpoints, two-session clothing variation, and 13M+ annotations across detection, tracking, ReID, and action recognition, \emph{DetReIDX} is the first dataset to comprehensively reflect the constraints of long-range UAV-based pedestrian ReID.

Our benchmarks show that state-of-the-art detectors and ReID models degrade their performance up to 81\% when tested on the \emph{DetReIDX} set. Also, models still face particular difficulties in case \emph{within-subject} cloth changes, which is a fundamental requirement for long-term ReID. Hence, \emph{DetReIDX} should not be regarded as a simple convenience benchmark, but - instead - as a stress test and a foundation tool. It shall set a new standard for evaluating the robustness of models and a challenge to support the development of real-world models.

\section*{Acknowledgment}

Kailash A. Hambarde acknowledges that this work was carried out within the scope of the project “Laboratório Associado”, reference CEECINSTLA/00034/2022, funded by FCT – Fundação para a Ciência e a Tecnologia, under the Scientific Employment Stimulus Program. The author also thanks the Instituto de Telecomunicações for hosting the research and supporting its execution.

Hugo Proença acknowledges funding from FCT/MEC through national funds and co-funded by the FEDER—PT2020 partnership agreement under the projects UIDB/50008/2020 and POCI-01-0247-FEDER-033395.

\bibliographystyle{IEEEtran}
\bibliography{references}

\end{document}